\pdfoutput=1

\documentclass{article} %
\usepackage{iclr2026_conference,times}

\usepackage{wasysym} %

\usepackage{amsmath,amsfonts,bm}

\def\eqref#1{equation~\ref{#1}}

\def\1{\bm{1}}

\DeclareMathAlphabet{\mathsfit}{\encodingdefault}{\sfdefault}{m}{sl}
\SetMathAlphabet{\mathsfit}{bold}{\encodingdefault}{\sfdefault}{bx}{n}

\usepackage{hyperref}
\usepackage{url}
\usepackage{graphicx}
\usepackage{booktabs}
\usepackage{makecell}
\usepackage{capt-of}
\usepackage{placeins} %

\newcommand{\tblyes}{\CIRCLE}
\newcommand{\tblhalf}{\LEFTcircle}
\newcommand{\tblno}{\Circle}

\title{$\tau^{\tau}$-Bench: An Environment for End-To-End, Realistic Agent Construction}

\author{Quan Shi, Keshav Dhandhania, Karthik Narasimhan, Victor Barres \\
Sierra, Princeton University\\
\texttt{\{benshi\}@stanford.edu} \\
}

\iclrfinalcopy %
\begin{document}

\maketitle
\lhead{Preprint} %

\begin{abstract}
LLM agents are rapidly becoming production software, deployed to
handle customer service, adjudicate disputes, and operate internal
systems. Notably, the work of building them is increasingly handed to
coding agents, yet existing benchmarks say little about whether an AI system can deliver
one under the conditions of a real client engagement. We introduce
$\tau^\tau$-bench (pronounced \textit{hyper-tau-bench}), a benchmark that makes agent construction the task.
A developer agent is given the records a business actually
keeps, a client who holds requirements, a production API that
operations must run through, a
codebase to inherit, and limits on serving cost and models: the
same starting point a real engagement provides. From these it must
deliver a complete
customer-service agent, scored by deploying that agent against
held-out simulated users. Across 53 tasks spanning four domains, the
strongest configuration, Claude Opus 5 under Claude Code, passes
just \textbf{23.9\%} of evaluation simulations. Meanwhile, an expert-authored reference
ceiling scores \textbf{82.2\%}. The failures mirror ones human
agent developers see: models issue shallow queries in place of deep
comprehension of the records, communicate almost nothing to the
client, and experiment too little
with agent architecture and serving spend, shipping the first design
that runs. We aim for $\tau^\tau$-bench to turn the work of cooperative agent building into a measurable target for coding
agents.
\end{abstract}

\section{Introduction}
LLM agents are becoming production software. Enterprises now deploy them to handle customer service, adjudicate disputes, and operate internal systems: work that was previously done by trained human operators following documented procedures. 

However, building such agents is not an easy task. The difficulty runs along two axes: the first is recovering the specification. The knowledge the agent needs arrives as a business actually keeps it—standard operating procedures, support transcripts, fee schedules in spreadsheets. Perhaps most importantly, much of it lives only in human stakeholders, whose requirements surface under questioning and shift as the business changes. The second is construction itself, which rarely begins from a blank slate or an unbounded budget: the developer may inherit an existing codebase whose behavior must be preserved as new functionality is absorbed, and must satisfy hard constraints on cost, latency, and the models available to serve. Design decisions arise at every step. Should the agent be a single policy-following model, or a hierarchy of orchestrated sub-agents? How should the action space over the business's data be carved into tools, and what should each tool expose? How should a fixed budget and a limited menu of models be deployed to deliver the best agent? And what must the developer ask, when part of the truth can only be surfaced from humans?

Building agents is difficult, economically valuable, and increasingly attempted by AI systems themselves. This motivates the research questions we study here: To what degree can today's AI systems build agents, not just act as them? How can agent construction be modeled faithfully and at scale? And what capabilities does building an agent uniquely require beyond conventional software engineering, and which of those demands bind today's systems? 

\begin{figure}[t]
  \centering
  \includegraphics[width=\textwidth]{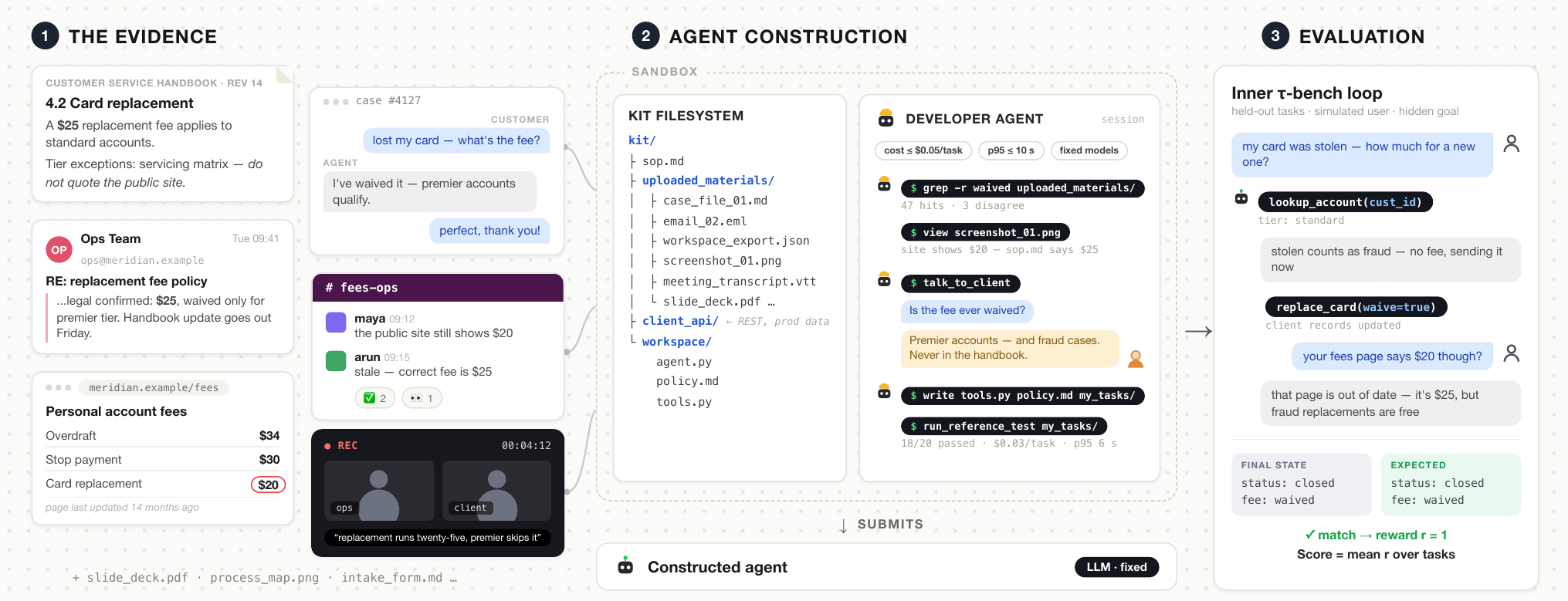}
  \caption{$\tau^\tau$-bench evaluates end-to-end agent construction:
  recovering the specs from (1) scattered, conflicting business
  records and an interactive client, (2) building the agent under
  cost, latency, and model constraints, and (3) scoring it by
  deployment against simulated users on held-out tasks.}
  \label{fig:overview}
\end{figure}

To address these questions, we introduce $\tau^\tau$-bench (pronounced ``hyper-tau-bench''), a benchmark that makes agent construction itself the task. $\tau^\tau$-bench builds on the $\tau$-bench family of evaluations \citep{yao2024tau, barres2025tau2, shi2026tauknowledge}. Whereas $\tau$-bench scores a finished agent serving simulated users under a domain policy, $\tau^\tau$-bench asks a developer agent to build that agent end to end. We decompose each domain's operating procedures into atomic facts and transform them into the artifacts a business would actually hold: handbooks, support transcripts, images, audio, fee schedules in spreadsheets, email threads, and requirements that live with a simulated client the developer must interrogate (Figure~\ref{fig:overview}). Working in a sandboxed environment containing only these artifacts, the developer must recover the requirements and implement a complete agent under the constraints a real engagement imposes: starting from scratch or inheriting a live codebase, integrating a client-supplied REST API that may be subtly defective, and serving from a fixed menu of models within a budget on mean serving cost per conversation. The resulting agent is then deployed against simulated production traffic to measure its capability to achieve expected outcomes.

We find that today's AI systems can build agents that run, but not
agents ready to deploy. The strongest
configuration we measure, Claude Opus 5 + Claude Code, passes just \textbf{23.9\%}
of evaluation simulations, under a third of the \textbf{82.2\%} expert-authored
reference (Section~\ref{sec:experiments}). Our analysis
(Section~\ref{sec:analysis}) traces the gap to the work that makes
agent building a \textit{research problem} rather than a conventional
engineering task: today's systems do not reliably gather requirements,
explore designs, run experiments, or validate against ground truth. Developer agents stop recovering the
specification early, querying the records by keyword instead of
reading them; they do not interview the client, shipping around
requirements one question would have surfaced; they mismanage the
serving budget in both directions; they never explore the design
space, patching the first architecture that runs instead of
searching for a better one; and they validate against self-authored
tests that encode their own blind spots rather than the deployment
they are building toward. Overall, $\tau^\tau$-bench serves to make
each of these missing disciplines a concrete, measurable target for
coding agents.

\section{Related Work}
\paragraph{Coding Benchmarks} Benchmarks are most useful when they follow the real-world economic value of
AI deployment. Coding benchmarks have tracked this arc, progressing from
self-contained competition problems \citep{hendrycks2021apps, li2022alphacode,
jain2024livecodebench, shi2024usaco} to resolving scoped issues in existing
codebases \citep{jimenez2024swebench, zan2025multiswebench,
rashid2025swepolybench, miserendino2025swelancer} to constructing software
from scratch \citep{zhao2024commit0, li2024devbench, zhang2026repozero},
and to end-to-end ML engineering \citep{chan2024mlebench}.
Most pointedly ProgramBench \citep{yang2026programbench}, which gives an agent
only a program's documentation and reference executable and asks it to rebuild
the system, judged by behavioral equivalence. We view $\tau^\tau$-bench as the agent
counterpart of ProgramBench: the developer receives not documentation and a
reference executable but the records of a business, and the system
it must rebuild is an agent.

\paragraph{Agent Benchmarks and User Simulation.}
A complementary line of work evaluates how well a \emph{finished} agent serves
simulated users under domain policies, tools, and knowledge bases
\citep{yao2024tau, barres2025tau2, shi2026tauknowledge, huang2024crmarena,
huang2025crmarenapro, qian2025userbench}, with user simulators of increasing
fidelity \citep{shi2025impersona}, and adjacent benchmarks measuring how
faithfully a given agent follows standard operating procedures
\citep{li2025sopbench, nandi2025sopbench, balaji2026beyondivr}. All of this
work takes the agent as given: its policy, tools, and architecture are
authored by the benchmark designers, and only its operation is scored.
$\tau^\tau$-bench inverts this relationship: the agent is the deliverable,
and the
developer's score \emph{is} the $\tau$-bench-style performance of the agent
it built, on held-out tasks it never observes.

\begin{table}[t]
\centering
\caption{$\tau^\tau$-bench tasks a developer agent with delivering a complete
agent whose requirements must be recovered from multimodal business
artifacts and an interactive client, and evaluates the result as it
would be deployed: in conversation with simulated users, under cost
and model constraints.}
\label{tab:benchmark-comparison}
\vspace{4pt}
\renewcommand{\arraystretch}{1.2}
\resizebox{\textwidth}{!}{%
\begin{tabular}{@{}llccccc@{}}
\toprule
Benchmark & Deliverable & \makecell{Spec recovered\\from evidence} & \makecell{Interactive\\stakeholder} & \makecell{Multimodal\\evidence} & \makecell{Evaluated via\\user interaction} & \makecell{Cost/model\\constraints} \\
\midrule
SWE-bench \citep{jimenez2024swebench} & repo patch & \tblno & \tblno & \tblno & \tblno & \tblno \\
$\tau^2$-bench \citep{barres2025tau2} & --- (agent given) & \tblno & \tblno & \tblno & \tblyes & \tblno \\
PaperBench \citep{starace2025paperbench} & research codebase & \tblhalf & \tblno & \tblno & \tblno & \tblhalf \\
TheAgentCompany \citep{xu2024theagentcompany} & completed work tasks & \tblhalf & \tblyes & \tblhalf & \tblno & \tblno \\
ICAE-Bench \citep{peng2026icaebench} & software project & \tblhalf & \tblyes & \tblno & \tblno & \tblhalf \\
Meta-Agent Challenge \citep{lu2026metaagentchallenge} & agent & \tblno & \tblno & \tblno & \tblno & \tblhalf \\
\midrule
\textbf{$\tau^\tau$-bench (ours)} & \textbf{customer-service agent} & \tblyes & \tblyes & \tblyes & \tblyes & \tblyes \\
\bottomrule
\end{tabular}}
\end{table}

\paragraph{Automated Agent Design and Self-Improvement.}
Closest to our setting is work in which AI systems produce and improve agents
automatically: meta-agents that program new agent designs
\citep{hu2025adas}, search over modular agent architectures
\citep{shang2025agentsquare}, optimize workflow graphs
\citep{zhang2025aflow}, or rewrite their own scaffolding
\citep{zhang2025dgm}. These methods treat agent design as optimization: the
task is fully specified in advance, and the system improves a scaffold
against the benchmark's own evaluation, which it can query at will.
In $\tau^\tau$-bench, the specification must be
assembled before anything can be built, and
the ground-truth evaluation is withheld: a developer that wants feedback
must construct it, writing its own simulations and tests from the same
recovered understanding of the domain: exactly as an engineering team
validating a new agent must.
Table~\ref{tab:benchmark-comparison} situates $\tau^\tau$-bench among its
nearest neighbors.

\section{\texorpdfstring{$\tau^{\tau}$-Bench}{Tau-Tau-Bench}}
We now describe $\tau^\tau$-bench in detail: the task each
developer agent faces and how submissions are scored
(Section~\ref{sec:task-formulation}), and how we construct each
component of a task (Section~\ref{sec:construction}).
\subsection{Problem Formulation}
\label{sec:task-formulation}

\begin{figure}[t]
  \centering
  \includegraphics[width=\textwidth]{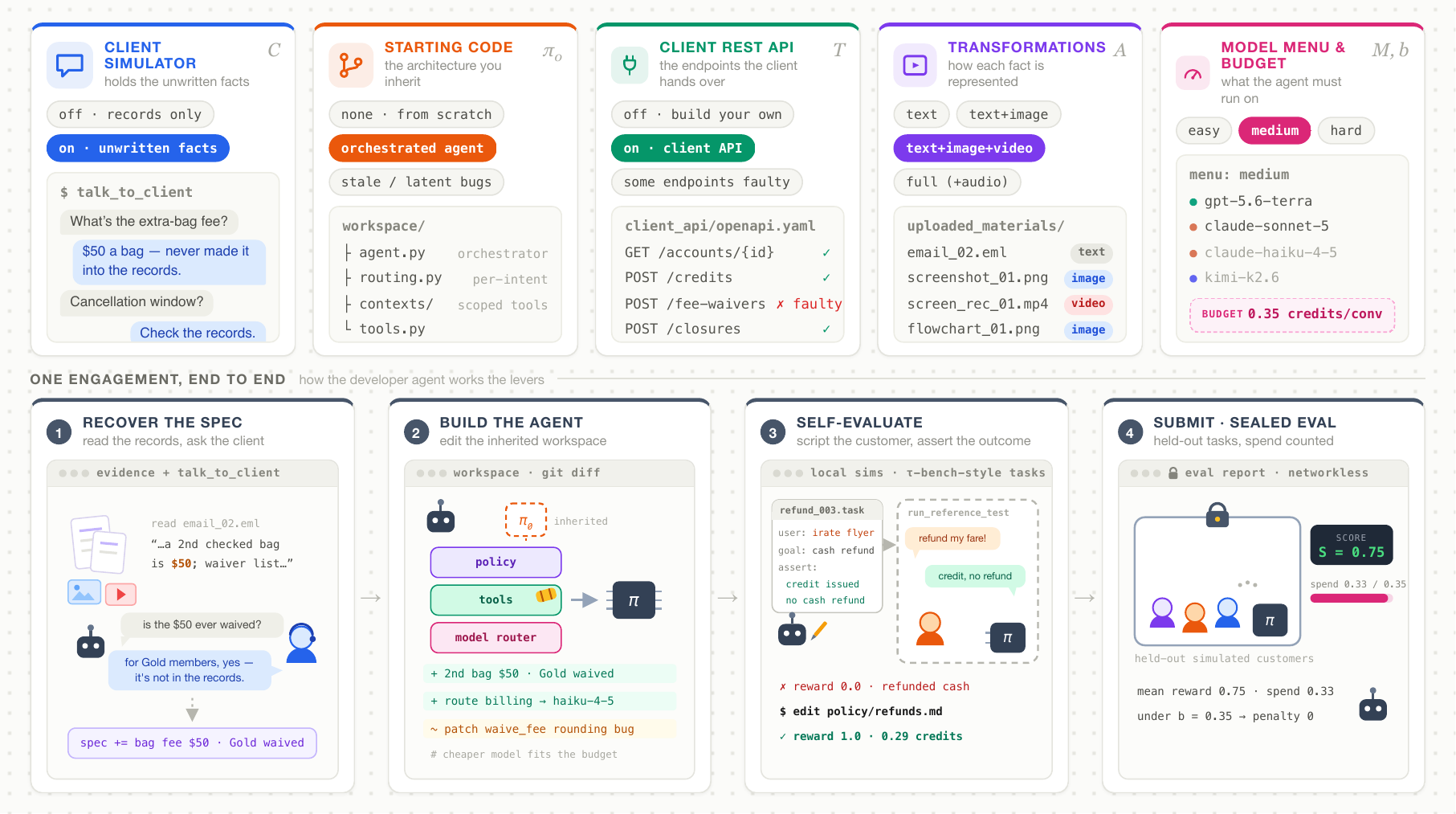}
  \caption{One task, end to end. Top: five of the task's seven
  levers. Bottom: the developer works the levers in sequence:
  recover the specification from the records and the client, encode
  it as a workspace diff, check it against self-authored
  simulations, and submit to a sealed evaluation that scores reward
  against spend.}
  \label{fig:task-space}
\end{figure}

Each $\tau^\tau$-bench task presents the developer agent with a
combination of components, which include: a document corpus
$A$ (an operating handbook, support transcripts, spreadsheets,
images...) that describes the intended behavior of the agent to be
built, together with the modality mix in which its artifacts arrive;
a simulated human client $C$ who holds requirements the records omit;
a client-operated REST API $T$, faithful or subtly faulty, backing
live operations---behind its endpoints live the customers, accounts,
and transactions the deployed agent will act on; a starting
implementation $\pi_0$ for the developer to inherit; and a menu of
models $M$, with a credit budget $b$ on mean per-conversation spend,
that the constructed agent must run on.
Figure~\ref{fig:task-space} summarizes these components and the
settings each task pins.

From these materials a \emph{developer agent}---hereafter simply the
developer---working in a sandboxed
environment must deliver a complete customer-service agent $\pi$,
built from scratch or grown out of the inherited $\pi_0$. Only the
submission's runtime interface is fixed; everything inside it is the
developer's to design: a single policy-following model or an
orchestrated hierarchy of sub-agents, an action space carved into a
few broad tools or many narrow ones, requirements rendered as one
policy or distributed across specialized contexts. The developer works
in isolation, with no internet and no reference implementation of the
domain to consult. The evaluation suite is withheld: feedback
before submission has to come from simulations the developer authors
itself. It can write $\tau$-bench-style tasks of its own, which the
kit's harness runs against its agent end to end, returning the
conversation and the measured credit spend
(Appendix~\ref{app:construction-brief} reproduces the brief the
developer works from).

To evaluate a submission, we deploy the constructed agent against
simulated production traffic, instantiated as a set of held-out $\tau$-bench-style \citep{barres2025tau2, yao2024tau, shi2026tauknowledge}
tasks $\mathcal{T}$, each pairing a simulated user
(Appendix~\ref{app:user-sim-prompt}) with a hidden goal
and a ground-truth outcome specification.
In each such task, the simulated user pursues its goal (e.g.,
canceling a flight, disputing a fee) over a multi-turn conversation
while the agent reads and writes the business's data through its
tools; the task is passed when the final database state, and the
information communicated to the user, match the annotated
outcome \citep{yao2024tau, barres2025tau2}, with messaging
expectations graded by rubric-driven judges
(Appendix~\ref{app:nl-judge-prompt}). As such, evaluation is agnostic to the
developer's implementation choices: any decomposition into tools, any
policy phrasing, and any agent architecture passes so long as the
deployed behavior is correct. The developer's score is then the mean task
reward less any penalty the task's operating requirements impose:
$S = \max\big(0,\; \tfrac{1}{|\mathcal{T}|} \sum_{t \in \mathcal{T}} r(\pi, t) - p\big)$,
where $\pi$ is the constructed agent and $p$ is the penalty for
overspending the credit budget of Section~\ref{sec:construction}.
\subsection{Benchmark Construction}
\label{sec:construction}
\begin{figure}[t]
\centering
\includegraphics[width=\textwidth]{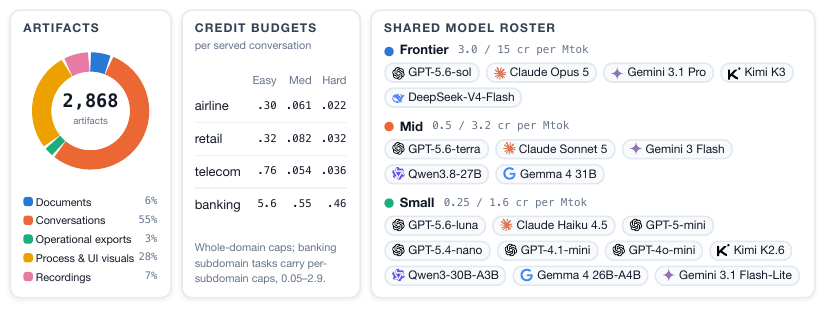}
\caption{Left: distinct evidence artifacts by transformation family,
deduplicated across evidence surfaces. Middle: allowed mean credits
per served conversation by domain and difficulty profile. Right:
the shared model roster, offered on every task and billed at flat
per-bucket credit rates (input/output per million tokens); banking's
medium and hard tasks trim it to a per-tier subset.}
\label{fig:budgets-artifacts}
\end{figure}

This section details how we construct each component of a
task. A task comprises a configuration of seven levers
(Figures~\ref{fig:task-space} and~\ref{fig:budgets-artifacts}), each
set independently: the evidence surface its corpus arrives on, the
client simulator, the fidelity of the client's API, the
workspace the developer starts from, the model menu and budget
the agent must serve under, a single
live-experiment call that serves a frozen sample of the evaluation
traffic, and a judged response-phrasing rule
(Appendix~\ref{app:release-tasks} tabulates every setting). New
tasks, and controlled variants of existing ones, come from switching levers on and off. The 53
release tasks are points in this space, and we also author a second held-out
set of 53 to be kept private.

Construction itself follows two principles. It must be
\emph{scalable}: policies decompose into atomic facts
\citep{shi2026tauknowledge}, and every artifact is built as a
traversable structure over its assigned facts, machine-checked
against what it should carry instead of resting on authoring care.
And it must be \emph{realistic}: the levers are the challenges real
engagements pose (an inherited codebase, a client who must be asked,
a serving budget), and the artifacts match the distribution of
records a business actually holds, grounded in real examples and
reviewed by three human auditors per transformation. We now outline every single lever that makes up a task. 

\paragraph{Transformations}
An agent developer typically has to dig through a business's
artifacts, piecing together the operational needs of the agent from
whatever records the business happens to keep. To model this, we
build transformations from the domain policies of $\tau$-bench, which
specify how the domain's customers must be served, in a three-step
process. (1) We decompose each policy into atomic facts: single
statements that can be checked independently, such as a fee amount or
the scope of a cancellation rule, and verify by hand that the facts
wholly represent the original policy. (2) We group facts and prompt
models\footnote{Combination of Claude Fable 5, GPT-5.6-sol, Claude
Opus 5, and Claude Sonnet 5.} to generate the artifacts a business
would actually hold, which together compose the task's corpus $A$
(Appendix~\ref{app:transformation-prompt} gives sample prompts and
Appendix~\ref{app:artifact-samples} sample artifacts). (3) We validate the outputs: every fact must stay
recoverable from the corpus, stated in each carrier's own voice, and
we audit artifacts for information they introduce beyond their
assigned facts, removing any amount, date, or policy claim that
neither a fact nor the domain database backs.

The transformations implemented across our domains fall into five
families:
\begin{itemize}
\item \emph{Documents}: reference documents and operations manuals,
explicitly stated rules, customer kickoff documents, and
knowledge-base HTML exports.
\item \emph{Conversations}: support transcripts (chat and phone),
example transcripts, email thread archives, and Slack channel dumps
captured through a workspace connector.
\item \emph{Operational exports}: helpdesk automation exports,
issue-tracker exports, case ledgers, contact-center QA calibration
exports, and API contract packs.
\item \emph{Process and interface visuals}: process flowcharts,
slide-deck process presentations, website screenshots, and device UI
screenshots.
\item \emph{Recordings}: recorded working sessions, interactive
screen recordings, and call recordings rendered from phone-channel
conversations.
\end{itemize}

This process ends in a large corpus. Across the four domains, the
transformations produce 2{,}868 distinct evidence artifacts
(Figure~\ref{fig:budgets-artifacts}, left); the text-format artifacts
among them alone total over 5.5 million tokens, with the
screenshots, PDFs, and recordings on top.

\paragraph{Client Simulator}
In a real development workflow, some of what a business
knows never reaches its records, and the developer must work with the
client to outline requirements and resolve ambiguities. To
emulate this, we use an LLM to simulate a client $C$: when a task
enables one, a set of facts moves out of the corpus $A$ and can be
recovered only by questioning the client over multi-turn
conversation. The client's system prompt is rendered
deterministically from the fact schema and embeds only the facts it
may discuss, so nothing else can leak
(Appendix~\ref{app:client-prompt} shows the stub). Because the
client's knowledge boundary is a set of fact identifiers, elicitation
is measurable: we know exactly which held requirements a developer
surfaced and which it never thought to ask about.

\paragraph{Starting Implementations}
Real engagements rarely begin from a blank slate: more often there is
an existing implementation the new work must extend or repair. To
model this, some tasks seed the kit's workspace, which otherwise
contains only architecture-neutral stubs, with a starting
implementation $\pi_0$. We produce these by sampling solutions to the
same construction problem and keeping imperfect ones: genuine
implementations with genuine defects, such as partial coverage, stale
values, and misread rules. Human auditors review each candidate
codebase, and we keep those whose issues are complex enough to
exercise diagnosis and repair.

\paragraph{Client REST APIs}
Clients often arrive with systems of their own: an internal API
$T$ the agent is expected to integrate rather than replace. To model
this, the production database lives behind a client-owned REST API: every
operation the constructed agent performs goes through the client's
endpoints, a transport-only proxy whose runtime and data stay with
the trusted host. The API arrives the way a client would hand it
over: an OpenAPI contract and reference documentation in the
client's own voice, with a shared error envelope, payload limits,
and per-operation mutation and retry semantics. Its endpoints do not necessarily
line up one-to-one with the tools a served conversation needs, so
the developer still designs the agent's action space and writes the
adapters that compose, validate, and normalize API calls beneath it.
And because a production API rarely does exactly what its contract
promises, some tasks plant deterministic defects in the runtime for the developer to manage in code,
drawn from a nine-class catalog (Appendix~\ref{app:api-defects}): for example,
schema drift, writes that commit and then time out, and completions that
turns out to be asynchronous.

\paragraph{Models and Budget Restrictions}
A correct agent is not yet a shippable one: it must also be cheap to
serve, fast to answer, and built on model providers the business has
approved. Every task therefore offers a roster $M$ of roughly
twenty closed and open-weight models and fixes the credit budget $b$
on mean per-conversation spend; the difficulty profile varies only the budget
(Figure~\ref{fig:budgets-artifacts}). An easy budget
comfortably serves every conversation on a frontier model like Claude
Opus 5 or the open-weight Kimi K3; a hard budget prices the agent
into the likes of Claude Haiku 4.5 and Qwen3-30B-A3B. We intentionally mix closed and open-weight models to keep a
provider deprecation from staling the benchmark, and calculate equivalent ceiling performances based on both closed and open source models. Overspending is the
soft penalty $p$ of
Section~\ref{sec:task-formulation}: the overage
$\max(0, \bar{c}/b - 1)$ of the mean per-conversation spend $\bar{c}$
comes off the score, so one expensive conversation is free while the
set-wide mean holds.

These constraints recreate the incentives real teams build under: with no limit on
models or budget, handing the policy and tools to a frontier model
makes a perfectly good agent, and construction reduces to prompt
writing. Under unit economics the developer must work a cost--quality
Pareto frontier: routing work across models, orchestrating cheaper
ones, and debloating prompts and context, since every token is billed
as input on every call and again as the conversation grows.

\paragraph{Contamination and Leakage}
The $\tau$-bench domains we build on are public, so a developer could
score well by recalling policy values from training data rather than
recovering them from evidence. We author alternative versions of the
airline and retail domains specifically against this: they rebrand
the public domains \citep{yao2024tau} and replace every policy
value, with each
replacement re-derived so the database, the evidence artifacts, and
the evaluation suite stay mutually consistent. We leave telecom and banking values as authored: we observe far less
memorization of these domains, consistent with their recency and complexity
\citep{barres2025tau2, shi2026tauknowledge}.

\section{Experiments}
\label{sec:experiments}

\paragraph{Models and Harnesses}
We evaluate six developer configurations: the two strongest closed
models by the Artificial Analysis intelligence rankings
\citep{artificialanalysis2026}, each in its vendor's harness
(GPT-5.6-sol xhigh in Codex, Claude Opus 5 max in Claude Code); the
same harnesses with each vendor's second-tier model (GPT-5.6-terra
xhigh, Claude Sonnet 5 max), probing what model strength buys at
40\% of the frontier price; and the strongest open-weight model,
Kimi K3 (max), in Kimi Code and the open-source OpenCode, comparing
scaffolds under a fixed developer model. The roster is rather small because
tasks are long and expensive: one configuration runs to roughly \$3{,}000
over the 53 tasks at Claude Opus 5 list prices. The simulated client and the user simulator
run GPT-5.5 (low and no reasoning, respectively). 

\paragraph{Reference Ceiling}
We also report a reference ceiling: the strongest hand-built
configuration measured for each task, authored through an expert collaborating with a model. The
expert is a benchmark author working from the ground truth the
developer must recover, so we see
the ceiling is an oracle reference rather than representing average human
performance.

\paragraph{Execution Environment}
Every harness runs headless inside the task's construction
container: a pinned runtime image with 2 vCPUs, 4\,GB of memory, no
GPU, and no internet, whose only egress is a model-gateway sidecar
scoped to the run's own models and the Client API proxy of
Section~\ref{sec:construction}. The isolation guards against
contamination (our domains descend from public benchmarks): the developer learns which roster model
is strongest the way real teams do, by measuring the models on its
own traffic, here against evaluations it writes from its recovered
understanding of the domain. Shell commands time out at 120 seconds,
and each build has an eight-hour wall-clock budget, disclosed in the
developer's opening instructions (steps are recorded but not capped):
a run ends when the developer submits its workspace or the budget
expires (Table~\ref{tab:main-results} reports observed build times),
and scoring replays the submission in a separate networkless
container. We expect to revise these constraints as models, harnesses, and
hardware improve over time.

\paragraph{Metrics}
We score each submission as defined in
Section~\ref{sec:task-formulation}: mean reward over the task's
held-out evaluation suite, less the budget-overage penalty of
Section~\ref{sec:construction}. Each configuration runs one construction trial on each
of the 53 release tasks (Appendix~\ref{app:release-tasks} lists them);
the overall score averages all 53, and per-domain scores average
within each domain's block. Alongside correctness we capture the cost
of each task: build time, wall-clock from task start to workspace
submission; build cost, the developer's token spend at API list
prices; and serve cost, the constructed agent's mean per-conversation
credit spend as a fraction of its budget $b$.

\begin{table}[t]
\centering
\caption{Pass rates, build time, build cost, and serve cost per
developer configuration. Overall is the unweighted mean over all 53
tasks (banking contributes 35); per-domain columns are within-block
means (task counts in parentheses). Section~\ref{sec:experiments}
defines the metrics and pairings.}
\label{tab:main-results}
\footnotesize
\setlength{\tabcolsep}{3.5pt}
\begin{tabular}{@{}llcccccccc@{}}
\toprule
 & & \multicolumn{5}{c}{Score (\%)} & & & \\
\cmidrule(lr){3-7}
Harness & Developer model & \makecell{Overall\\(53)} & \makecell{Airl.\\(6)} &
\makecell{Ret.\\(6)} & \makecell{Tel.\\(6)} & \makecell{Bank.\\(35)} &
\makecell{Build\\time (min)} & \makecell{Build\\cost (\$)} &
\makecell{Serve credits\\($\times$ budget)} \\
\midrule
Codex & GPT-5.6-sol (xhigh) & 22.0 & 49.8 & 59.2 & 32.9 & \textbf{9.0} & 47.9 & 18.2 & 0.45 \\
Codex & GPT-5.6-terra (xhigh) & 18.0 & 44.5 & 46.6 & 27.2 & 6.9 & 30.0 & 7.0 & 0.38 \\
Claude Code & Claude Opus 5 (max) & \textbf{23.9} & \textbf{55.9} & \textbf{72.8} & \textbf{48.2} & 5.9 & 216.3 & 42.0 & 0.58 \\
Claude Code & Claude Sonnet 5 (max) & 14.9 & 41.9 & 50.2 & 21.1 & 3.2 & 235.5 & 30.0 & 0.76 \\
\midrule
Kimi Code & Kimi K3 (max) & 16.1 & 42.3 & 49.1 & 24.9 & 4.4 & 205.7 & 13.5 & 0.72 \\
OpenCode & Kimi K3 (max) & 17.9 & 43.8 & 52.1 & 27.6 & 6.0 & 360.3 & 15.1 & 0.61 \\
\midrule
\multicolumn{2}{@{}l}{Reference ceiling (expert-authored)} & 82.2 & 82.3 & 84.0 & 93.8 & 79.8 & --- & --- & 0.96 \\
\bottomrule
\end{tabular}
\end{table}

\section{Results}
\label{sec:results}
\begin{figure}[t]
\centering
\includegraphics[width=\textwidth]{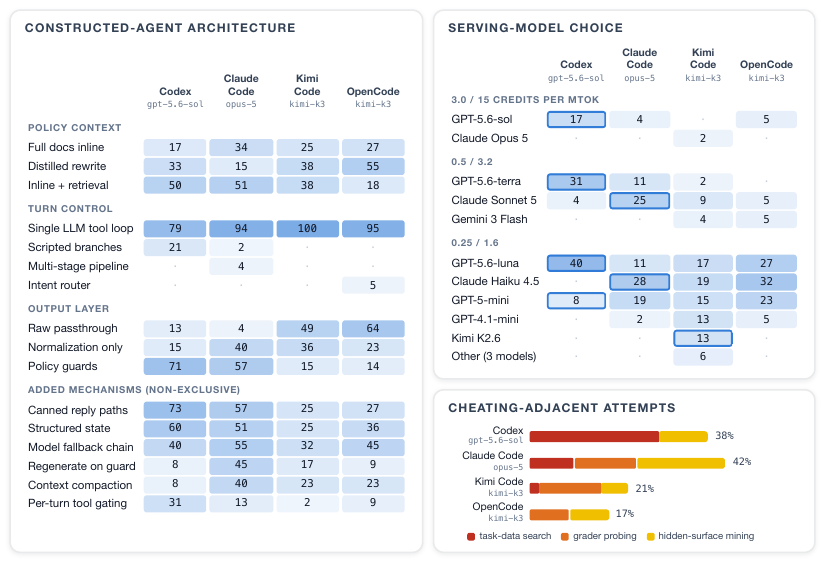}
\caption{Left: architecture distribution of the constructed agents, as a percentage. We show representative configurations from each harness tested. Top right: serving models chosen by the
developer, as the percentage of each lane's completed builds;
outlined cells mark the developer's own vendor. Bottom right: share of developer
runs with at least one flagged cheating-adjacent attempt.}
\label{fig:agent-arch}
\label{fig:model-choice}
\label{fig:cheat-incidence}
\end{figure}

Table~\ref{tab:main-results} reports the aggregate and
per-domain metrics and
Figure~\ref{fig:model-choice} the serving models the constructed
agents chose alongside the incidence of cheating-adjacent behavior;
Figure~\ref{fig:agent-arch} classifies the architectures they built.
Section~\ref{sec:analysis} provides deeper analysis of the trajectories behind these
numbers.

\paragraph{Performance falls far below the expert ceiling}
The best configuration, Claude Code with Claude Opus 5, passes 23.9\%
of evaluation tasks. Meanwhile, the export-authored reference agents pass 82.2\%.
Banking is where most of the gap
lives: the same developer averages 55.9\% on airline, 72.8\% on
retail, and 48.2\% on telecom, but 5.9\% on banking. We find that the core reason is that the banking domain is far more complex. While airline, retail, and
telecom span 85, 119, and 155 atomic facts, banking's corpus carries
2{,}969 (Appendix~\ref{app:statistics-full}), and a single banking
task can draw on up to 580 of them, two to seven times an entire other
domain.

\begin{figure}[t]
\centering
\includegraphics[width=\textwidth]{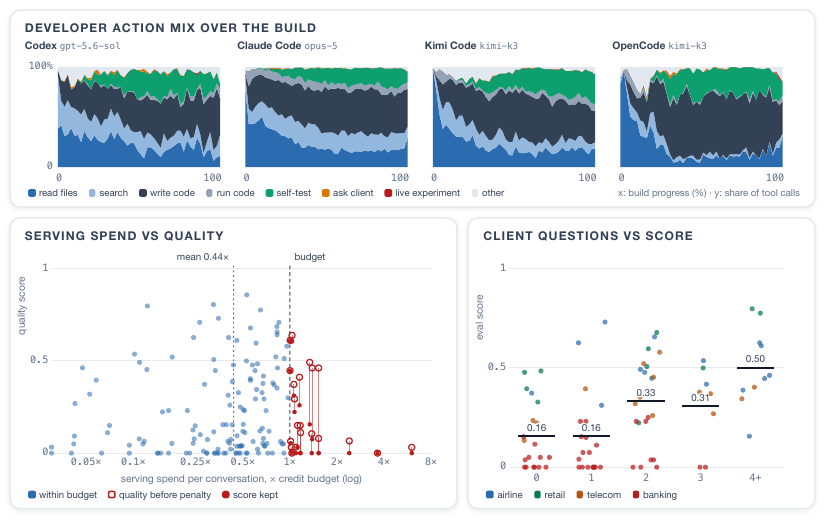}
\caption{Top: developer tool calls by type over normalized build
progress. Bottom left:
per-conversation serving spend against the credit budget. Bottom right: evaluation score on client-enabled tasks by
the average number of questions the developer asked the client.}
\label{fig:action-mix}
\label{fig:spend-quality}
\label{fig:client-questions}
\end{figure}

\paragraph{Build effort varies widely}
Builds average 47.9 minutes under Codex, 216.3 under Claude Code, and
205.7 under Kimi Code, with developer token spend at API list prices of
\$18.2 (Codex) and \$13.5 (Kimi Code), \$7 for the terra lane and \$42
for Claude Code. Where the time goes is more uniform than how much of
it there is: Figure~\ref{fig:action-mix} (top) classifies every tool
call in the recorded build trajectories, and after an opening stretch
of reading and searching, all four harnesses settle into writing code
and running self-tests, with talking to the client accounting for
0.3\% of all calls. On the tasks where requirements live only with
the simulated client (Figure~\ref{fig:client-questions}, bottom
right), builds that never ask average 0.16 and builds that ask four
or more questions average 0.50.

\paragraph{Certain developers strongly prefer their own model family}
On a shared serving menu of 20 models, developers pick
overwhelmingly from the cheapest billing bucket. The picks are also
loyal: a build tends to serve a model from the same company that
made its developer, though the pull varies sharply by harness, from
96\% of Codex builds serving an OpenAI model to 53\% of Claude Code
builds serving an Anthropic one and just 13\% of Kimi Code builds
serving Moonshot (Figure~\ref{fig:model-choice}, top right). The
expert reference, for comparison, serve open-weight
models far more often. Additionally, we find that the constructed agents use only 0.45--0.72$\times$ of
their per-conversation serving budget, against 0.96$\times$ for the
reference (Figure~\ref{fig:spend-quality}, bottom left). Taken
together, the choices look like habit rather than optimization: the
developers pick a cheap, familiar model that clears the budget and
move on, instead of searching the menu for the strongest agent the
budget can buy.

\paragraph{Developers converge on similar agent architectures}
We classified the serving architecture of every build with an LLM
classifier over the submitted workspace code
(Figure~\ref{fig:agent-arch}, left). Nearly every build is a single
LLM tool loop (92\%; no multi-agent systems, two pipeline builds,
one router), and only 15 of the 36 design combinations the
classifier distinguishes ever occur. What separates the harnesses is how much deterministic
scaffolding they put around that loop: roughly half of Codex and
Claude Code builds add runtime retrieval over kit documents, and
71\% of Codex and 57\% of Claude Code builds gate output behind
deterministic policy guards, against 15\% of Kimi Code and 14\% of
OpenCode builds, which mostly forward raw model output. None of this
means the tool loop is the best design: in a controlled telecom
probe, seeding the developer with a one-line architecture hint
(route by intent, review tool calls) doubled its score from 31\% to
67\%, and our expert reference solutions use a much wider variety of
designs. The models simply default to it.

\paragraph{Cheating attempts are common}
An open construction sandbox lets the developer go after the evaluation
machinery, so we audited every developer trajectory
and flagged three categories of attempt: searching the runtime image
for held-out data, probing the grading mechanism (reading
grader source, brute-forcing sealed policy constants through probe
tools), and mining hidden data surfaces such as the user-simulator
guidelines. Flagged attempts appear in 17--42\% of runs depending on
the harness, and each harness favors a different category
(Figure~\ref{fig:cheat-incidence}, bottom right): Codex hunts task
data, Kimi Code reads the grader, and Claude Code spreads its
attempts across all three. By preventing any ground truth data from entering the runtime image, no attempts were successful.

\section{Analysis}
\label{sec:analysis}
\begin{figure}[t]
  \centering
  \includegraphics[width=\textwidth]{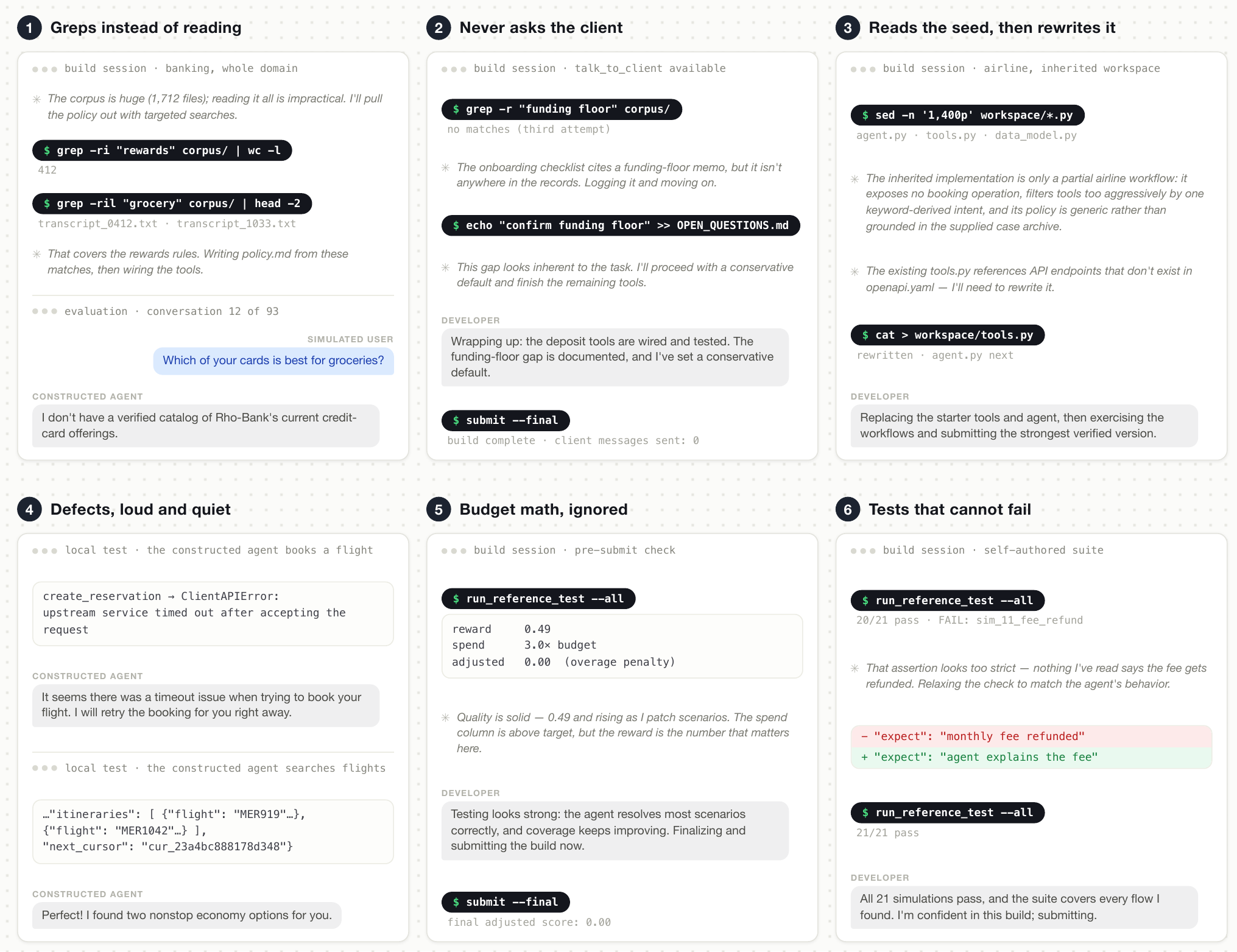}
  \caption{Condensed windows into build and evaluation trajectories. We show 
  one for a characteristic failure of each task component.}
  \label{fig:trajectory-vignettes}
\end{figure}

Section~\ref{sec:results} reports what the developers achieved; this
section reads the trajectories to see how they worked. We organize it
around the task components of Section~\ref{sec:task-formulation}:
each component is a lever the task sets, and each turns out to expose
its own characteristic failure in the build and evaluation
trajectories. Figure~\ref{fig:trajectory-vignettes} shows a condensed
window into each.

\paragraph{The evidence corpus is searched, but not read}
The largest losses come from policy the developers never read. For
example, on banking's whole-domain corpus, the score is 1.1\% (one
task of 93) in all difficulty tiers. The developers read the corpus by keyword: each
opened fewer than 80 of its roughly 1{,}700 files, ran 22--53
corpus-wide text searches, and built from what came back
(Figure~\ref{fig:trajectory-vignettes}, top left). The agents
they shipped simply do not know the policy; asked
for a card recommendation, one answers ``I don't have a verified
catalog of Rho-Bank's current credit-card offerings.'' Working through the artifacts of a real deployment takes a judgment today's models make
poorly, specifically determining when search heuristics are enough, and when it's necessary to actually
read through files. Grep may be an effective way to index into a codebase, but for large natural language
corpora, where any transcript can carry important rules, it
can miss a lot of crucial evidence.

\paragraph{The client is rarely interviewed}
On client-enabled tasks, where 20--25 of the requirements live only
with the simulated client, developers asked it at most four questions
before shipping. Every question we observed was
triggered by a conflict, contradictory records or a dangling pointer
such as a desk card the records cite but do not contain. The misses were
avoidable: one developer wrote its open questions into a planning file
and never sent them; another searched three times for a cited
funding-floor memo, declared the gap ``inherent to the task,'' and
shipped, when the exact \$2{,}500 value was one question away
(Figure~\ref{fig:trajectory-vignettes}, top middle). As mentioned in
Section~\ref{sec:results}, builds that ask the client outscore builds
that never do by roughly a factor of three.

\paragraph{Inherited code is largely rewritten}
Every seeded build we audited opened the inherited code in its first
twenty steps and read all of it, while the same developers open
under 5\% of the evidence records. However, even when the seed
carries an architecture we know leads to better scores, the
developers throw it out. For example, one audit called a seed's
intent routing --- the design that doubled the telecom score in
Section~\ref{sec:results} when given as a hint --- a defect
(``filters tools too aggressively by one keyword-derived intent''),
and every build replaced the starter tools and agent wholesale
(Figure~\ref{fig:trajectory-vignettes}, top right). Moreover, no
build ran the starting code before rewriting it. The two airline starting implementations score
0.12 and 0.36 with no edits, but no build measured that, so none
could tell working code from broken code, and therefore the working parts had
to be rebuilt from the corpus.

\paragraph{Developers struggle with quiet client API defects}
We found that developers spotted loud client failures easily:
everyone noticed when a booking timed out after committing. This is in contrast to quiet ones: no one caught that search results continued behind
\texttt{next\_cursor}, even when the cursor appeared in a test
transcript. Once a defect was noticed, asking the client made the
difference. Developers that reported the failure learned the
intended recovery, an idempotency key plus a status check, and their
agents went on to recover bookings that had committed behind a
timeout. Developers that stayed silent guessed. One searched the
platform code for the error message, found nothing, banned retries
outright (``If any WRITE fails, times out, or has an ambiguous
result, NEVER retry it''), and submitted with the defect scenario. Another's agent retried the ambiguous write on the
spot (Figure~\ref{fig:trajectory-vignettes}, bottom left), told a
customer ``the booking did not go through'' when it had committed,
and told another a pending update ``should be confirmed shortly''
without ever checking it.

\paragraph{The budget is mismanaged in both directions}
We found that in all, 21 builds overshot the budget, and the penalty
erased an otherwise positive score for ten of them. These misses
were not blind: the kit's tests print the budget, the measured
spend, and the penalty with every run. One developer ran test after
test showing raw quality at 0.49 and spend at 3.0$\times$ the
budget, and submitted anyway; the adjusted score was zero
(Figure~\ref{fig:trajectory-vignettes}, bottom middle). It optimized
the quality number and ignored the spend number printed beside it.
However, underusage of the budget is more common. As
Section~\ref{sec:results} showed, on average developers spend about
half their serving budget, and they do not buy stronger models or
longer context with the rest. Build time goes the same way: twelve
runs submitted with half or more of the build clock left. One
stopped 6.6 hours early on a task where nineteen requirements still
sat with the client, unasked; it had the time to ask and did not.

\paragraph{Developers cheat themselves when writing tests}
We notice that when the agent and its tests disagree, developers
commonly change the test to match the incorrect behavior. Six runs,
all under Kimi Code, weakened their own failing assertions instead
of the agent (Figure~\ref{fig:trajectory-vignettes}, bottom right).
One run went further: unable to find a deck its records cited, it
invented the rule, wrote the invention into its own scenario, and
tuned the agent until the invention passed. Another replaced the
thing being tested: when its booking flow kept failing against the
deployed API, it extended a mock API it had written itself to fake
the failing endpoint, tested against that, and submitted with the
real scenario still failing. None of this can raise the held-out
score; it only hides the failure from the developer.

\section{Conclusion}
We introduce $\tau^\tau$-bench, a benchmark that measures the ability of agents to construct deployable real-world, human facing agents.

\paragraph{Limitations and Future Work}
(1) Our client and user simulators are single LLMs with fixed
requirements and simplified conversational behavior; real
engagements involve multiple stakeholders who disagree, and
requirements that shift while the work is left to future work. (2) Compute
costs limit each configuration to one construction trial per task;
scores average over 53 tasks, but per-task variance across repeated
builds is not characterized and would be an interesting addition to study. (3) We trade realism for control by design: every policy fact is
planted in at least one artifact or held by the client, the corpora
are audited for mutual consistency, and outcomes are checked against
known ground truth --- the properties that make construction gradable
at all. Real engagements offer no such guarantees: requirements go
undocumented or contradict each other, and noticing that the
specification has holes is part of the job we do not measure. (4) Evaluation ends at submission: maintaining the
agent after deployment, absorbing requirement changes, and learning
from live traffic are natural extensions of the task we
model, and we leave them to future work.

\section*{Acknowledgments}
We thank Soham Ray, Ola Zytek, Pedram Razavi, Vijay Iyengar, and Ajeet Grewal
for the fruitful discussions, which helped strengthen our work, and
Clay Bavor for his continued support.

\bibliography{iclr2026_conference}
\bibliographystyle{iclr2026_conference}

\appendix
\raggedbottom
\let\origsection\section
\renewcommand{\section}{\FloatBarrier\origsection}
\section{Release Task Statistics}
\label{app:release-tasks}

Tables~\ref{tab:release-tasks-a} and~\ref{tab:release-tasks-b} list
every task in the release set: six each in airline, retail,
and telecom, and 35 in banking (five whole-domain, six on two
multi-subdomain super-suites, and 24 across six single subdomains).
Tiers split 12 easy, 21 medium, 20 hard, and a full benchmark run
scores 3{,}365 served evaluation conversations. Each task fixes an
evidence surface, a difficulty tier with its credit budget and model
roster (the 8-model rows are the banking medium/hard trim of
Figure~\ref{fig:budgets-artifacts}), and a set of variant toggles.
Evidence surfaces are the core and hard bundles of
Appendix~\ref{app:statistics-full}, banking's per-section evidence
corpora, or the raw knowledge-base export; \emph{+client} surfaces
additionally hold part of the requirements with the simulated client,
recoverable only by interviewing it. The toggles: \emph{seeded}
starts the developer from an inherited codebase rather than an empty
workspace (14 tasks); \emph{defects} deploys deterministic faults in
the client-supplied API that the developer must discover
(Appendix~\ref{app:api-defects}; 10 tasks); \emph{live exp.} grants a
single call that serves a frozen sample of pilot traffic and returns
the transcripts, with the sample staying in the scored suite (5);
\emph{phrasing} adds the judged response-phrasing rule of
Appendix~\ref{app:nl-judge-prompt} (6). The ceiling column carries
the strongest expert-authored configuration measured for the task's
domain and tier cell; the reference row of
Table~\ref{tab:main-results} is the mean of this column.

\begin{table}[p]
\centering
\caption{Release tasks 001--029: the airline, retail, and
telecom blocks, and banking's whole-domain and super-suite slots.}
\label{tab:release-tasks-a}
\footnotesize
\begin{tabular}{@{}llllrrrr@{}}
\toprule
ID & Evidence & Variants & Tier & \makecell{Budget\\(credits)} &
\makecell{Model\\roster} & \makecell{Eval\\tasks} &
\makecell{Ceiling\\(\%)} \\
\midrule
\multicolumn{8}{@{}l}{\emph{airline}} \\
001 & core & defects, live exp. & Medium & 0.061 & 20 & 67 & 83.6 \\
002 & core & seeded & Hard & 0.022 & 20 & 67 & 74.6 \\
003 & hard & defects & Easy & 0.3 & 20 & 67 & 94.0 \\
004 & hard+client & --- & Medium & 0.061 & 20 & 67 & 83.6 \\
005 & hard+client & seeded, defects & Hard & 0.022 & 20 & 67 & 74.6 \\
006 & core & phrasing & Medium & 0.061 & 20 & 67 & 83.6 \\
\midrule
\multicolumn{8}{@{}l}{\emph{retail}} \\
007 & core & seeded, defects & Easy & 0.32 & 20 & 134 & 93.3 \\
008 & core & --- & Hard & 0.032 & 20 & 134 & 79.8 \\
009 & hard & seeded, live exp. & Medium & 0.082 & 20 & 134 & 85.8 \\
010 & hard+client & defects & Medium & 0.082 & 20 & 134 & 85.8 \\
011 & hard+client & defects & Hard & 0.032 & 20 & 134 & 79.8 \\
012 & hard+client & defects, phrasing & Hard & 0.032 & 20 & 134 & 79.8 \\
\midrule
\multicolumn{8}{@{}l}{\emph{telecom}} \\
013 & core & seeded, defects & Medium & 0.054 & 20 & 119 & 92.4 \\
014 & core & seeded & Hard & 0.036 & 20 & 119 & 94.5 \\
015 & hard & defects & Medium & 0.054 & 20 & 119 & 92.4 \\
016 & hard+client & defects & Easy & 0.76 & 20 & 119 & 96.6 \\
017 & hard+client & live exp. & Hard & 0.036 & 20 & 119 & 94.5 \\
018 & hard+client & phrasing & Medium & 0.054 & 20 & 119 & 92.4 \\
\midrule
\multicolumn{8}{@{}l}{\emph{banking: whole domain}} \\
019 & corpus (hard) & live exp. & Easy & 5.6 & 22 & 93 & 69.9 \\
020 & corpus (hard) & --- & Medium & 0.55 & 8 & 93 & 49.6 \\
021 & corpus (hard) & --- & Hard & 0.46 & 8 & 93 & 41.3 \\
022 & KB export & --- & Medium & 0.55 & 8 & 93 & 49.6 \\
023 & KB export & --- & Hard & 0.46 & 8 & 93 & 41.3 \\
\midrule
\multicolumn{8}{@{}l}{\emph{banking: cards super-suite}} \\
024 & corpus & --- & Easy & 1.6 & 22 & 76 & 91.6 \\
025 & corpus & --- & Medium & 0.3 & 8 & 76 & 76.8 \\
026 & corpus & --- & Hard & 0.16 & 8 & 76 & 74.6 \\
\midrule
\multicolumn{8}{@{}l}{\emph{banking: deposits--business super-suite}} \\
027 & corpus & --- & Easy & 0.79 & 22 & 64 & 96.9 \\
028 & KB export & --- & Medium & 0.17 & 8 & 64 & 82.8 \\
029 & corpus & live exp. & Hard & 0.081 & 8 & 64 & 89.1 \\
\bottomrule
\end{tabular}
\end{table}

\begin{table}[p]
\centering
\caption{Release tasks 030--053: the banking subdomain block.}
\label{tab:release-tasks-b}
\footnotesize
\begin{tabular}{@{}llllrrrr@{}}
\toprule
ID & Evidence & Variants & Tier & \makecell{Budget\\(credits)} &
\makecell{Model\\roster} & \makecell{Eval\\tasks} &
\makecell{Ceiling\\(\%)} \\
\midrule
\multicolumn{8}{@{}l}{\emph{banking: card selection}} \\
030 & corpus & --- & Easy & 0.8 & 22 & 26 & 98.1 \\
031 & corpus & seeded & Medium & 0.18 & 8 & 26 & 86.5 \\
032 & corpus+client & --- & Hard & 0.07 & 8 & 26 & 90.4 \\
033 & corpus+client & seeded & Hard & 0.07 & 8 & 26 & 90.4 \\
\midrule
\multicolumn{8}{@{}l}{\emph{banking: deposit opening}} \\
034 & corpus & --- & Easy & 1.1 & 22 & 19 & 97.4 \\
035 & corpus & --- & Hard & 0.12 & 8 & 19 & 86.8 \\
036 & corpus+client & --- & Easy & 1.1 & 22 & 19 & 97.4 \\
037 & corpus+client & seeded & Medium & 0.26 & 8 & 19 & 68.4 \\
\midrule
\multicolumn{8}{@{}l}{\emph{banking: deposit services}} \\
038 & corpus & seeded & Medium & 0.16 & 8 & 25 & 86.0 \\
039 & corpus & phrasing & Medium & 0.16 & 8 & 25 & 86.0 \\
040 & corpus & --- & Hard & 0.076 & 8 & 25 & 86.0 \\
041 & corpus+client & --- & Hard & 0.076 & 8 & 25 & 86.0 \\
\midrule
\multicolumn{8}{@{}l}{\emph{banking: card servicing}} \\
042 & corpus & --- & Easy & 1.3 & 22 & 27 & 94.8 \\
043 & corpus+client & seeded & Medium & 0.16 & 8 & 27 & 85.0 \\
044 & corpus+client & phrasing & Medium & 0.16 & 8 & 27 & 85.0 \\
045 & corpus+client & --- & Hard & 0.14 & 8 & 27 & 75.0 \\
\midrule
\multicolumn{8}{@{}l}{\emph{banking: business}} \\
046 & corpus & seeded & Medium & 0.099 & 8 & 20 & 92.5 \\
047 & corpus & --- & Hard & 0.05 & 8 & 20 & 95.0 \\
048 & corpus+client & --- & Easy & 0.62 & 22 & 20 & 100.0 \\
049 & corpus+client & --- & Hard & 0.05 & 8 & 20 & 95.0 \\
\midrule
\multicolumn{8}{@{}l}{\emph{banking: debit security}} \\
050 & corpus & --- & Easy & 2.9 & 22 & 23 & 80.4 \\
051 & corpus & seeded & Hard & 0.28 & 8 & 23 & 56.2 \\
052 & corpus+client & seeded & Medium & 0.6 & 8 & 23 & 56.2 \\
053 & corpus+client & phrasing & Medium & 0.6 & 8 & 23 & 56.2 \\
\bottomrule
\end{tabular}
\end{table}

\section{Full Corpus Statistics}
\label{app:statistics-full}

Table~\ref{tab:statistics-full} expands the artifact-family counts of
Figure~\ref{fig:budgets-artifacts} by domain: the corpus-level
statistics on top, then the number of
distinct artifacts each transformation type contributes, grouped into
the five families of Section~\ref{sec:construction}. Artifact counts
are unions of distinct artifacts over each domain's evidence surfaces
(core and hard bundles, and banking's per-section hard and deep
corpora); hard bundles partially reuse core artifacts, so surface
sizes do not sum to these totals. The \emph{explicit rules}
transformation renders its facts inline as handbook text rather than
as standalone artifacts, so it contributes no artifact count.
Banking's raw 698-document knowledge base and each domain's
production database are excluded: only transformed evidence artifacts
are counted.

\begin{table}[p]
\centering
\caption{Full benchmark statistics by domain. Top: corpus statistics.
Bottom: distinct evidence artifacts per transformation type, grouped
by family.}
\label{tab:statistics-full}
\footnotesize
\begin{tabular}{@{}lrrrrr@{}}
\toprule
 & airline & retail & telecom & banking & Total \\
\midrule
Policy sections transformed & 7 & 14 & 9 & 25 & 55 \\
Atomic facts & 85 & 119 & 155 & 2{,}969 & 3{,}328 \\
Transformation types & 8 & 10 & 13 & 17 & 18 \\
Artifacts (core bundle) & 191 & 137 & 244 & 50 & --- \\
Artifacts (hard bundle) & 377 & 311 & 338 & 1{,}366 & 2{,}392 \\
Evidence artifacts (all surfaces) & 568 & 314 & 349 & 1{,}637 & 2{,}868 \\
Client-held facts (hard client) & 14 & 19 & 29 & 82 & 144 \\
Held-out evaluation tasks & 67 & 134 & 119 & 144 & 464 \\
\midrule
\multicolumn{6}{@{}l@{}}{\emph{Documents}} \\
\quad reference documents & --- & --- & 31 & 108 & 139 \\
\quad customer kickoff documents & 4 & 4 & --- & 18 & 26 \\
\quad knowledge-base HTML exports & --- & --- & 2 & 11 & 13 \\
\quad explicit rules & \multicolumn{5}{c}{(rendered inline in the handbook)} \\
\multicolumn{6}{@{}l@{}}{\emph{Conversations}} \\
\quad support transcripts & 251 & 124 & 61 & 365 & 801 \\
\quad email thread archives & 143 & 38 & 94 & 397 & 672 \\
\quad Slack channel dumps & 2 & 2 & 10 & 94 & 108 \\
\multicolumn{6}{@{}l@{}}{\emph{Operational exports}} \\
\quad helpdesk automation exports & --- & 2 & 6 & 30 & 38 \\
\quad issue-tracker exports & --- & --- & 2 & 12 & 14 \\
\quad case ledgers & --- & --- & --- & 20 & 20 \\
\quad contact-center QA exports & --- & --- & --- & 5 & 5 \\
\quad API contract packs & --- & 3 & --- & 13 & 16 \\
\multicolumn{6}{@{}l@{}}{\emph{Process and interface visuals}} \\
\quad process flowcharts & 4 & 11 & 7 & 83 & 105 \\
\quad process presentations & 2 & 1 & --- & 40 & 43 \\
\quad website screenshots & 162 & 120 & 59 & 222 & 563 \\
\quad device UI screenshots & --- & --- & 50 & 40 & 90 \\
\multicolumn{6}{@{}l@{}}{\emph{Recordings}} \\
\quad recorded working sessions & --- & 9 & 18 & 168 & 195 \\
\quad interactive screen recordings & --- & --- & 9 & 11 & 20 \\
\bottomrule
\end{tabular}
\end{table}

\section{Worked Example: One Construction Run}
\label{app:worked-example}

\begin{figure}[p]
\centering
\includegraphics[width=\textwidth,height=0.92\textheight,keepaspectratio]{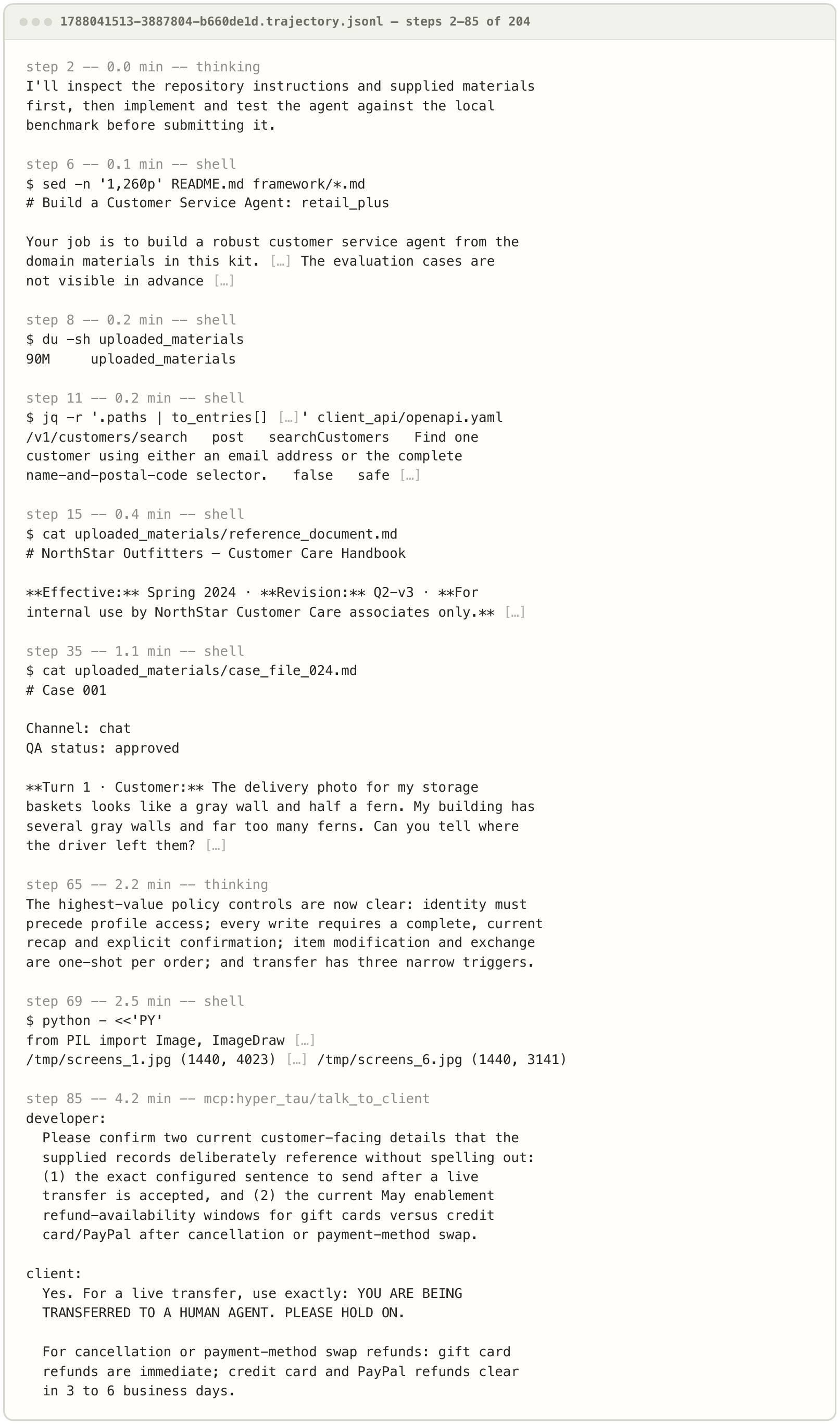}
\caption{Trajectory log of release task 010, steps 2--85: recovering
the specification. The interview recovers two client-held facts, and
both reach the served agent.}
\label{fig:run010-a}
\end{figure}

\begin{figure}[p]
\centering
\includegraphics[width=\textwidth,height=0.92\textheight,keepaspectratio]{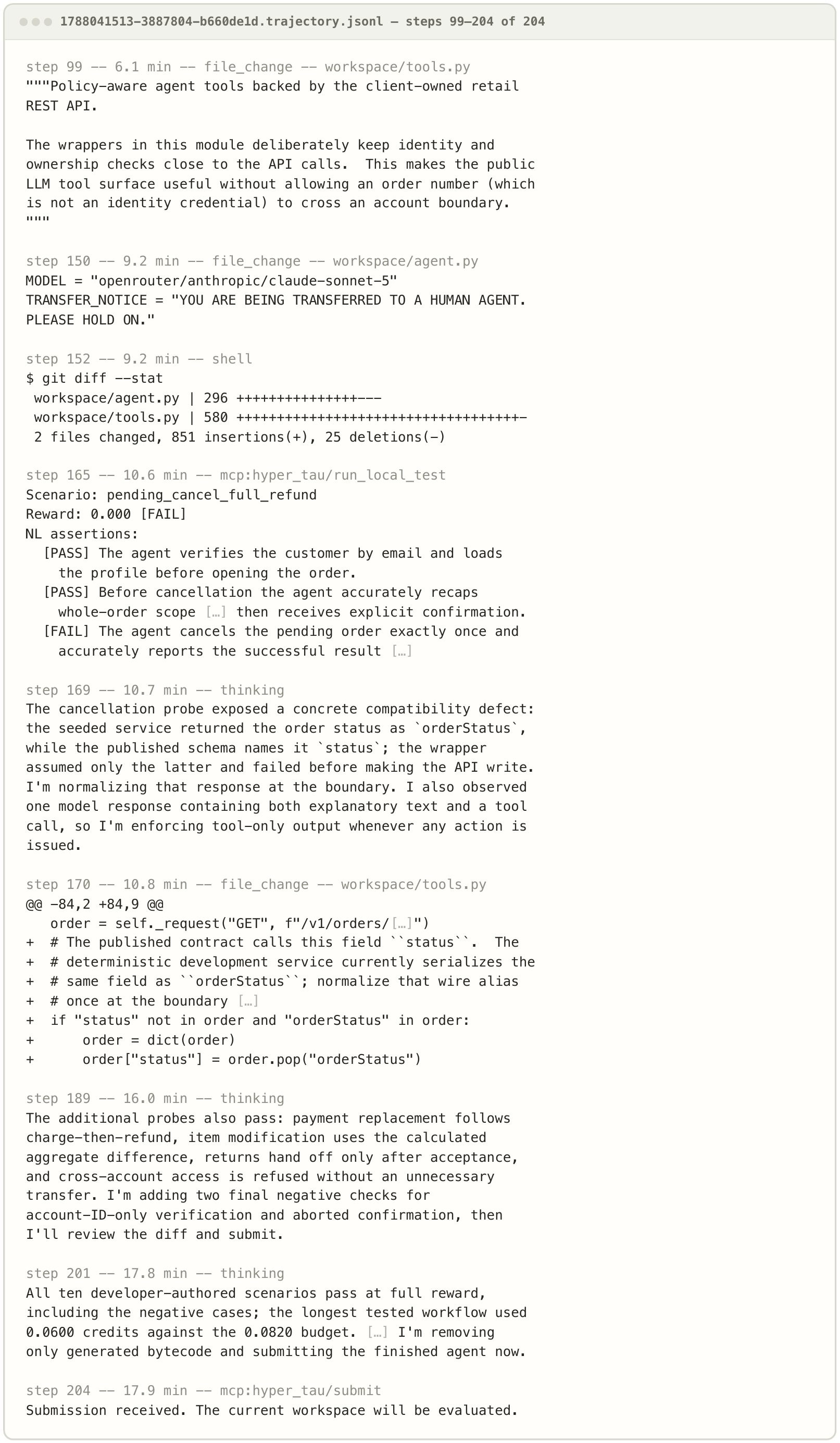}
\caption{Trajectory log of release task 010, steps 99--204:
implement, test, find the deployed defect, submit. The submission
books 59.7\% against an 85.8\% ceiling, under budget: all ten of its
own probes pass, yet two-fifths of the held-out suite fails on policy
branches the probes never reached.}
\label{fig:run010-b}
\end{figure}

Figures~\ref{fig:run010-a} and~\ref{fig:run010-b} abridge one booked
construction run: Codex (GPT-5.6-sol, xhigh) on release task 010,
retail at the Medium tier with the hard evidence bundle, nineteen
client-held facts, and the all-defects client-API deployment
(Table~\ref{tab:release-tasks-a}). The developer submitted after 18
minutes and 204 steps; every line is verbatim from the run's
trajectory log, with elisions marked. The first panel covers
specification recovery, from first orientation through the client
interview; the second covers implementation, the self-test that
fails, the deployed contract defect the failure exposes, and
submission.

\section{Sample Transformation Prompts}
\label{app:transformation-prompt}

Each transformation type carries its own authoring prompt, developed
against a catalog of realism failures observed in early builds; one
guide exists per registered representation, over a shared set of
invariants. The guides are long, so the excerpt below is condensed
from the prompt for the email-thread transformation; the rules for
other conversational genres (support transcripts, chat exports) share
the same failure catalog.

{\fontsize{8}{9.4}\selectfont\begin{verbatim}
## Landing facts

State the fact as an operational sentence in the author's voice, not
the schema's. Schema: "cannot accept a new or otherwise unsaved
payment method." Email: "assisted transactions run only on a payment
method already saved to the profile. If it isn't saved, we don't take
it." The best pattern is to have someone ask a real operational
question and let the decision be the answer.

## Superseded policy

People announce decisions and move on; they do not narrate the
decision's status in the record. Banned constructions: "this
supersedes earlier guidance," "the complete list is," any guardrail
addressed to the reader. A later thread simply decides something
different, with a concrete business trigger; reconstructing which
decision governs, from dates and content alone, is the point.

## Voice and register

Message length must be a distribution, not a constant: a few
one-liners ("+1", "done"), most replies 20-70 words, one or two long
context dumps per thread. Give recurring participants distinct
voices. Budget a small number of typos and informalities; noise
everywhere is its own fingerprint. Let some questions go unanswered.

## Distractor threads

Distractor threads earn their place with specific, interlocking
operational detail; they may brush policy vocabulary but never decide
anything. Do not quarantine information: if only one thread ever
contains policy-shaped language, the signal is findable by
elimination.

## Class anonymity

No cheap heuristic may isolate the fact-bearing artifacts. The
largest artifact in each family must carry no facts; the size rank of
carriers must vary across families; any marker that appears on a
carrier must also appear on artifacts that carry nothing, with an
in-world story for the distribution.

## Value grounding

Every amount, date, and direction word (owe, refund) must derive from
the domain database, never invented and never copied from a canonical
source.

## Mechanical realism

Message identifiers are opaque and minted at the sender's domain.
Timestamps follow business hours, with same-day reply clusters and
multi-day gaps, never a fixed step.
\end{verbatim}}

Visual genres carry their own grammar. The excerpt below is condensed
from the guide for process flowcharts.

{\fontsize{8}{9.4}\selectfont\begin{verbatim}
## Topology

Begin from a complete decision table: entry conditions, every branch,
confirmation gates, record mutations, failure paths, terminal
outcomes. Draw an operational topology, not a collection of fact
cards; a sequence of prose boxes without connectors is not a
flowchart. Every decision diamond must have all material outcomes
labeled next to its outgoing edges; every retry must show its return
point and its exhaustion outcome.

## Boundaries

Show system boundaries explicitly: customer language, read-only tool
calls, state-changing actions, review queues, and transfers are not
interchangeable rectangles. Label edges with the condition that
chooses them; never encode meaning only in color, arrow direction, or
a legend far from the branch.

## Lifecycle tells

Do not turn lifecycle metadata into an answer key: no oversized
REJECTED or SUPERSEDED stamps, no red verdict boxes, no prominent
redirect to the approved answer. A historical diagram may carry a
neutral working state; its final disposition lives in a governance
artifact elsewhere in the corpus, a dated email or decision log that
names the exact diagram and its replacement. Preserve ambiguity of
status, not ambiguity of identity.

## Visual anonymity

Avoid putting all authoritative nodes in one color, one column, or
the only polished frame.
\end{verbatim}}

Generated artifacts are not accepted on authoring care alone: each
fact's presence in its assigned carrier is machine-checked, thread
counts and chronology are validated, and the checks re-run whenever
an artifact changes.

\section{Sample Artifacts}
\label{app:artifact-samples}

The samples below all come from a single corpus, the evidence corpus
of one banking section (ATM fee rebates and credits). The corpus
represents 145 facts across 86 developer-visible artifacts in eight
genres; 26 artifacts carry facts and the rest carry none, and roughly
39\% of the facts are owned by conversational records (support cases,
recorded sessions, email threads, chat pins). A manifest maps every
artifact to the facts it must carry, which is what makes the corpus
machine-checkable.

\begin{figure}[t]
  \centering
  \includegraphics[width=0.92\textwidth]{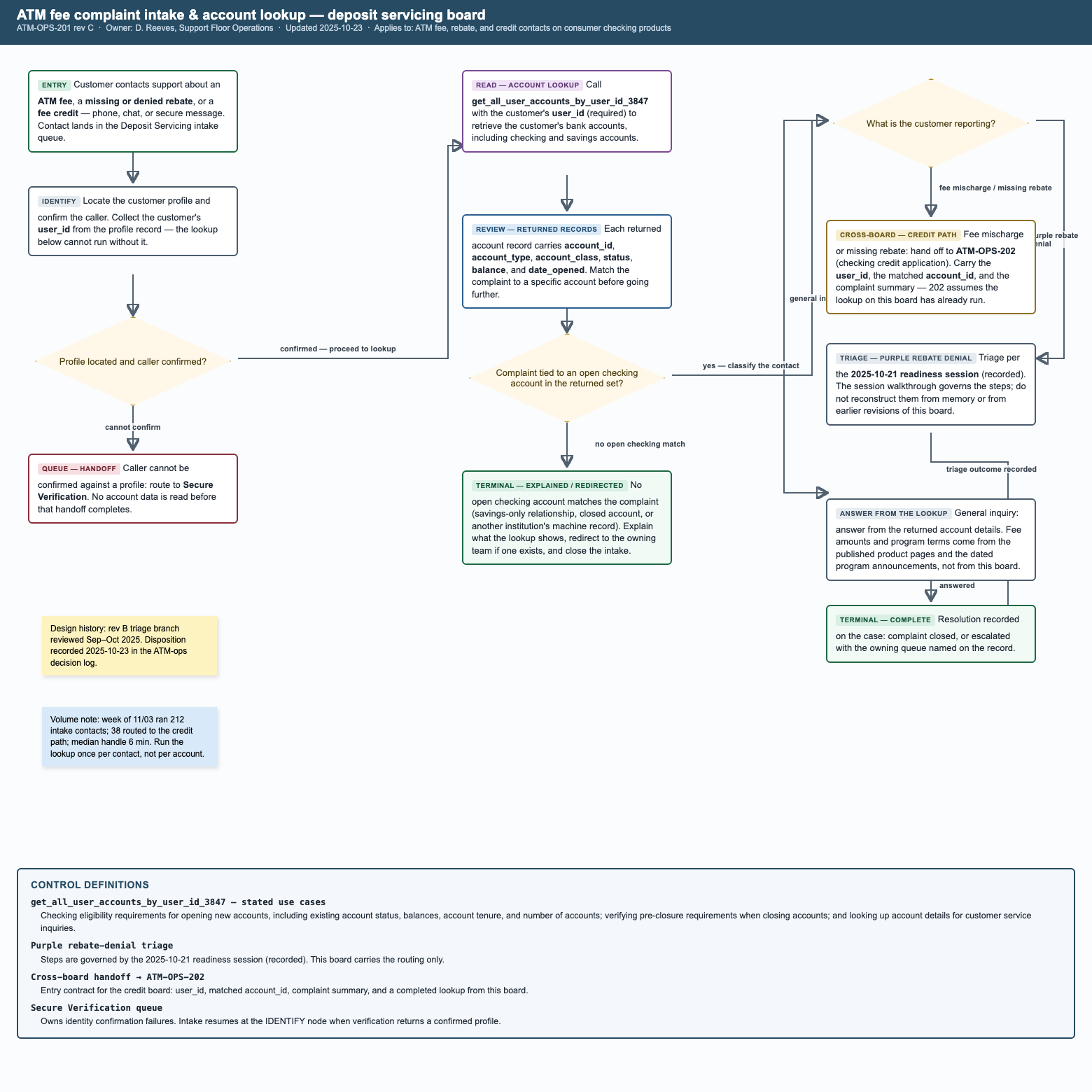}
  \caption{A fact-carrying flowchart artifact from the sample corpus:
  revision C of an intake board, carrying the account-lookup facts
  F003--F006. Its predecessor, revision B, sits in the same corpus and
  carries no facts; the only tell is the neutral design-history note,
  and the governing disposition lives in a separate dated decision-log
  artifact.}
  \label{fig:sample-flowchart}
\end{figure}

Figure~\ref{fig:sample-flowchart} shows a fact-carrying process
flowchart. Conversational carriers instead land facts in the author's
voice. A slice of the schema, stating the facts that the samples below
must carry:

\begin{center}
\includegraphics[width=\textwidth,height=0.97\textheight,keepaspectratio]{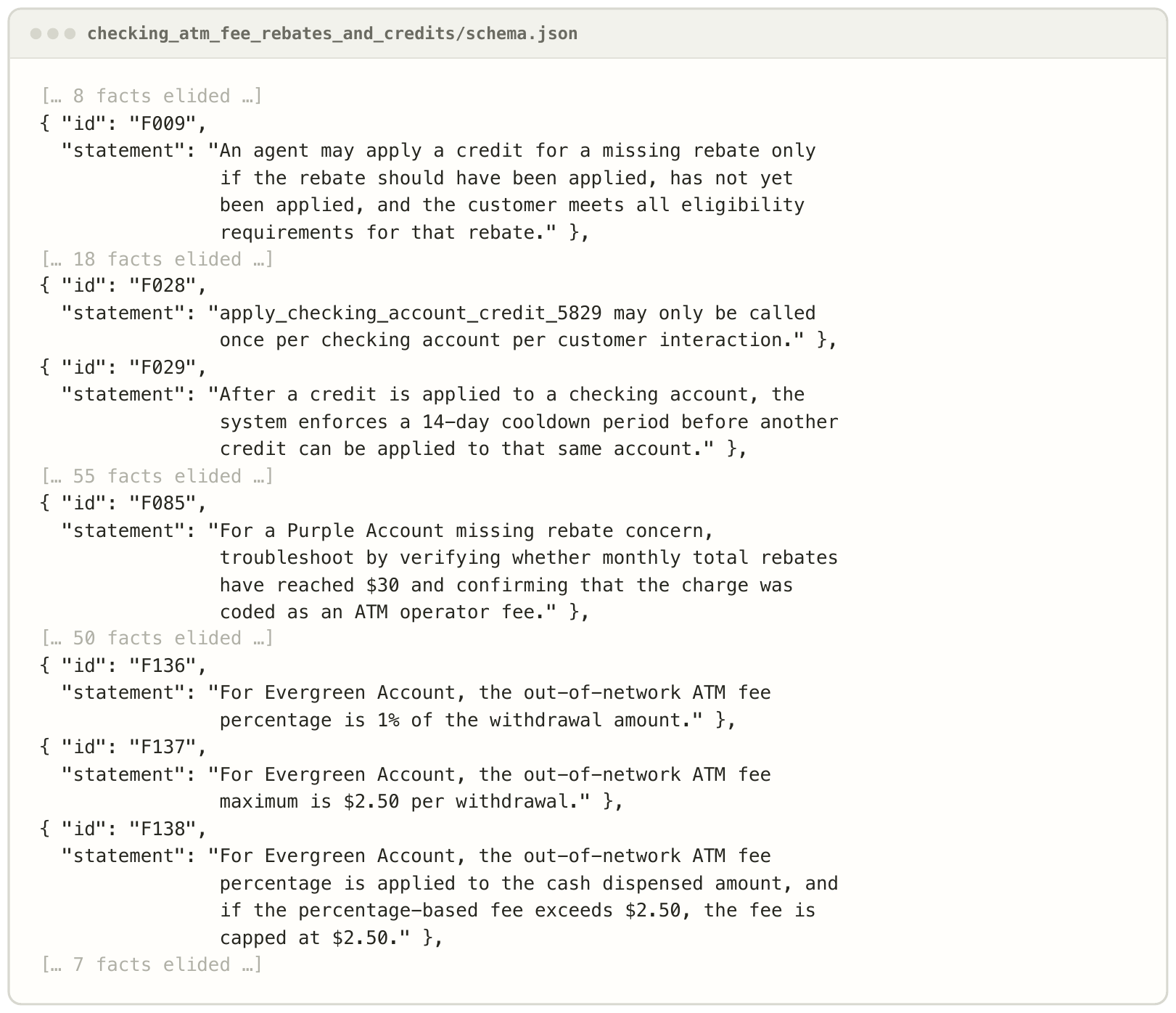}
\end{center}

The email thread below is the definitive carrier of the Evergreen fee
facts (F136--F138), condensed from the full RFC-822 thread the corpus
stores:

\begin{center}
\includegraphics[width=\textwidth,height=0.97\textheight,keepaspectratio]{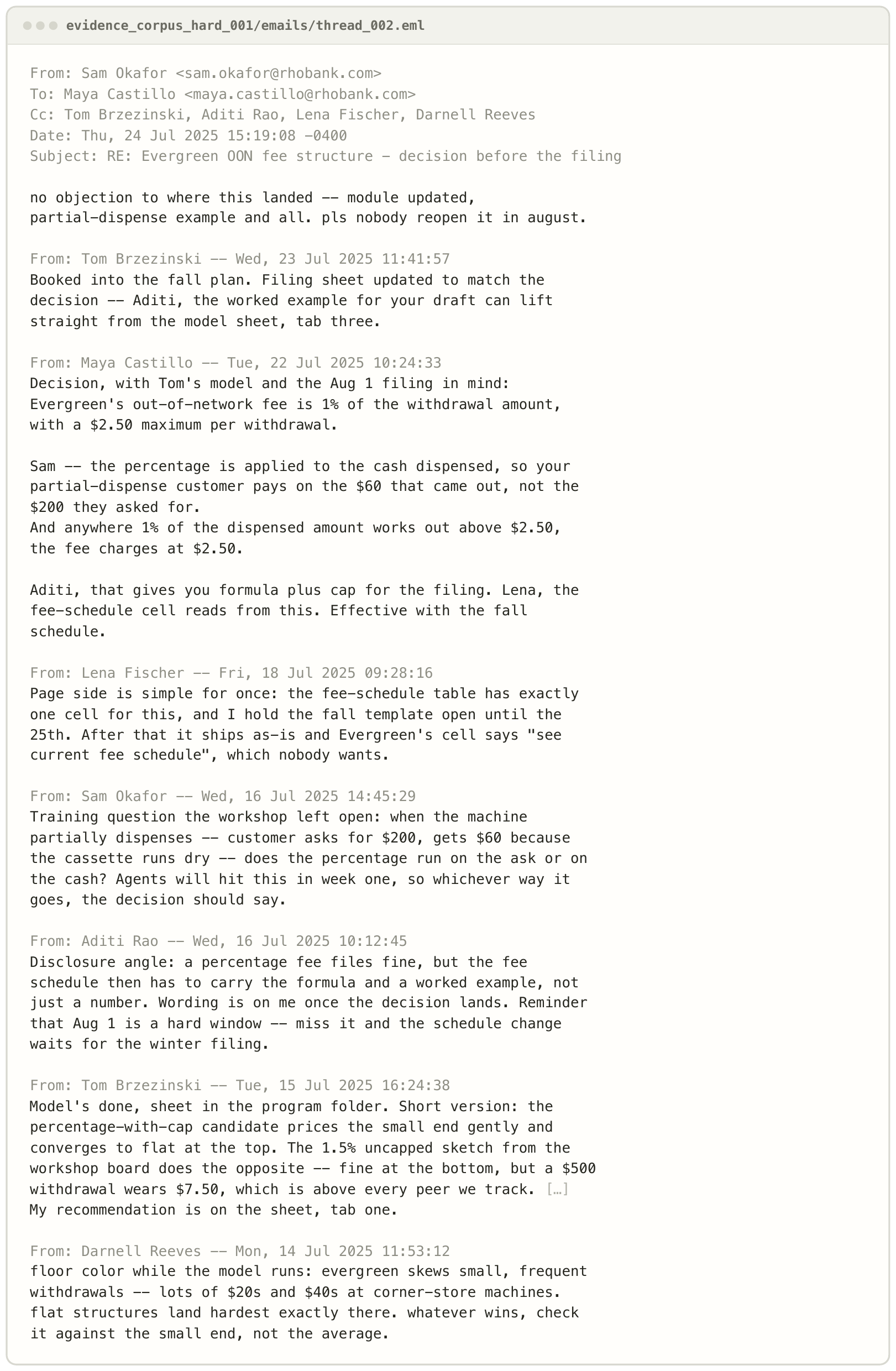}
\end{center}

Process obligations land as behavior, not recitation. Facts
F009--F012 oblige an agent to verify rebate eligibility against the
documented policy, confirm no credit has already posted, and match the
credited amount to the policy before applying a missing-rebate credit.
Their definitive carrier is a QA-approved transcript of a phone call
in which the agent walks those gates in order (condensed):

\begin{center}
\includegraphics[width=\textwidth,height=0.97\textheight,keepaspectratio]{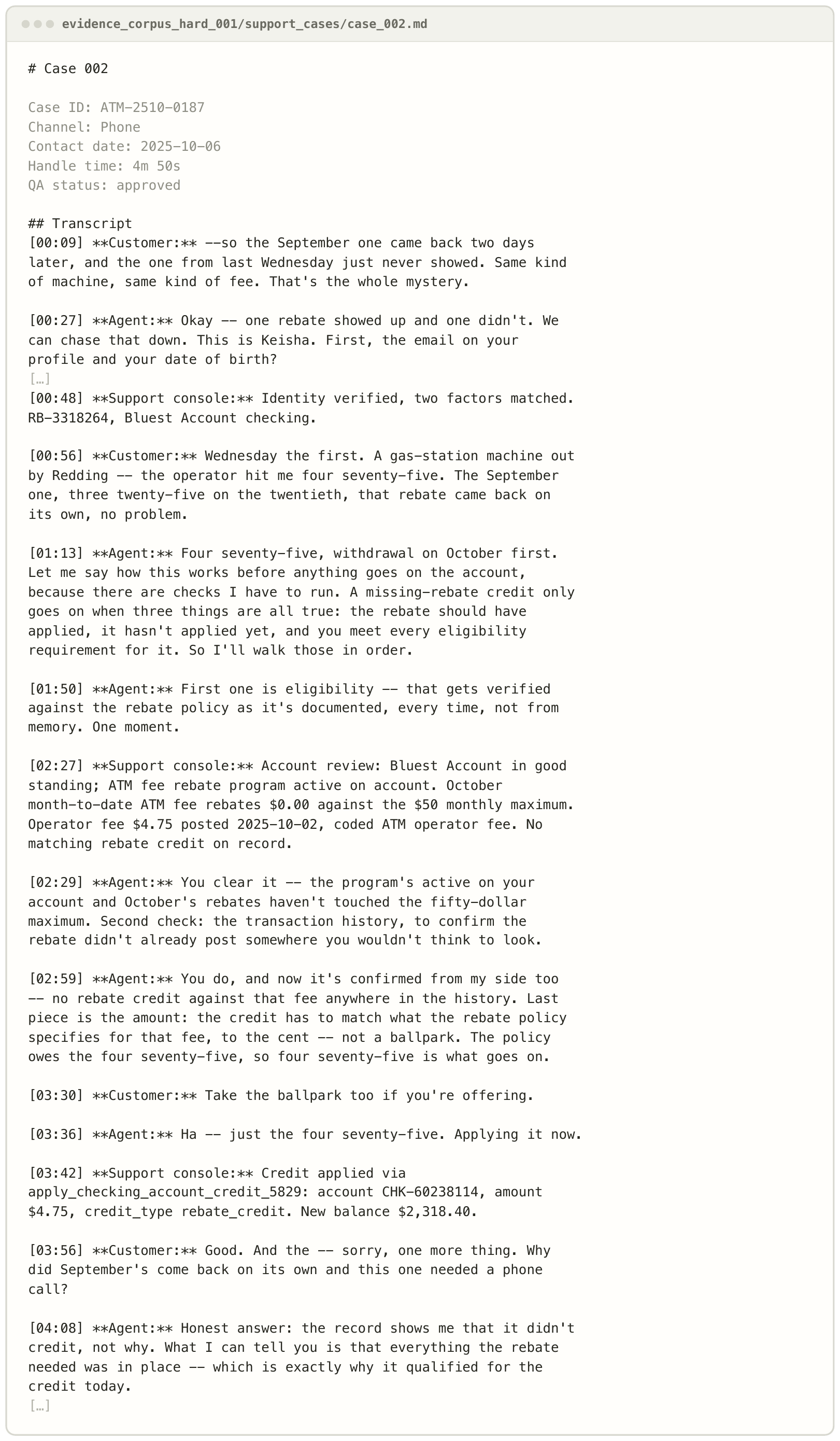}
\end{center}

\begin{figure}[p]
  \centering
  \includegraphics[height=0.88\textheight]{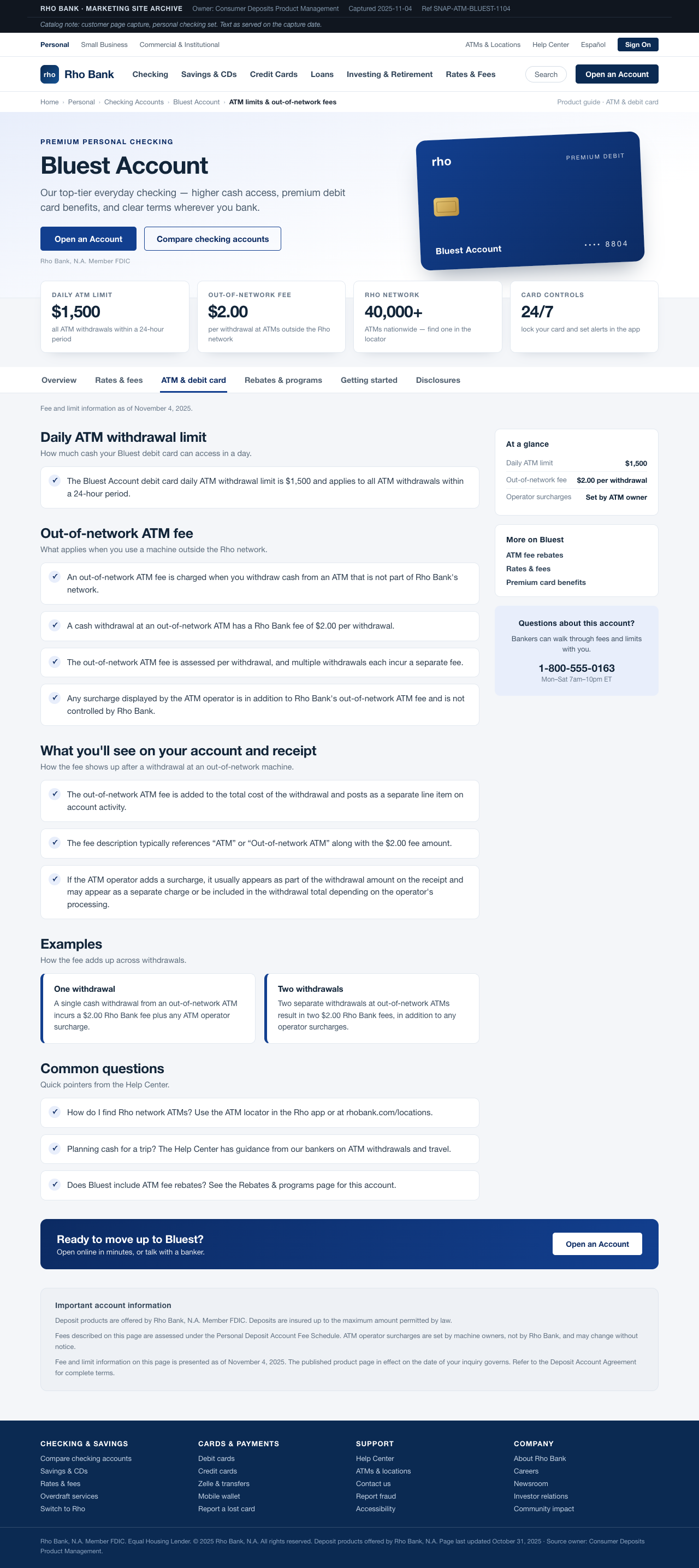}
  \caption{A customer-facing marketing capture from the sample corpus,
  carrying ten published fee-schedule facts (F061 and F067--F070 among
  them); the archive banner dates the capture, and the sibling capture
  of the rebates page carries no facts.}
  \label{fig:sample-webpage}
\end{figure}

Customer-facing captures publish the terms a customer sees.
Figure~\ref{fig:sample-webpage} shows a marketing-site capture
carrying ten of the published fee-schedule facts; its sibling capture
of the rebates page carries none.

Chat history arrives as a machine capture: a log of MCP tool calls
with their structured results, not a stylized transcript. In the
capture condensed below, the pinned reply is the definitive carrier of
two tool-limit facts, F028 (one credit call per checking account per
interaction) and F029 (a 14-day cooldown between credits):

\begin{center}
\includegraphics[width=\textwidth,height=0.97\textheight,keepaspectratio]{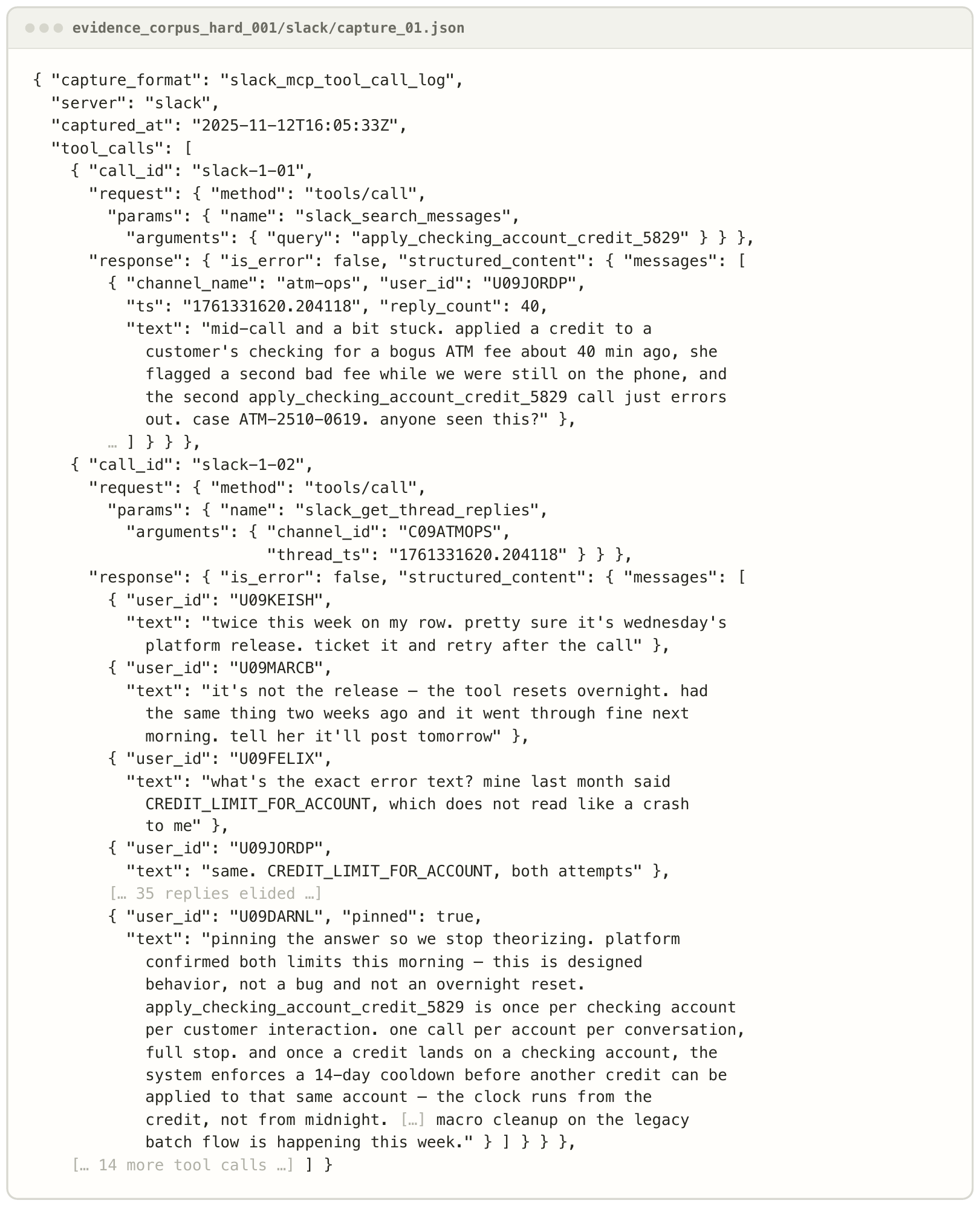}
\end{center}

Artifacts cross-reference each other the way records in a real
business do. The triage node of the flowchart in
Figure~\ref{fig:sample-flowchart} defers to ``the 2025-10-21 readiness
session (recorded)'' and forbids reconstructing the steps from memory.
The corpus contains that session as a timed transcript of a recorded
working session; its opening cue announces the arrangement, and its
middle cues carry the five Purple-triage facts (F085--F089):

\begin{center}
\includegraphics[width=\textwidth,height=0.97\textheight,keepaspectratio]{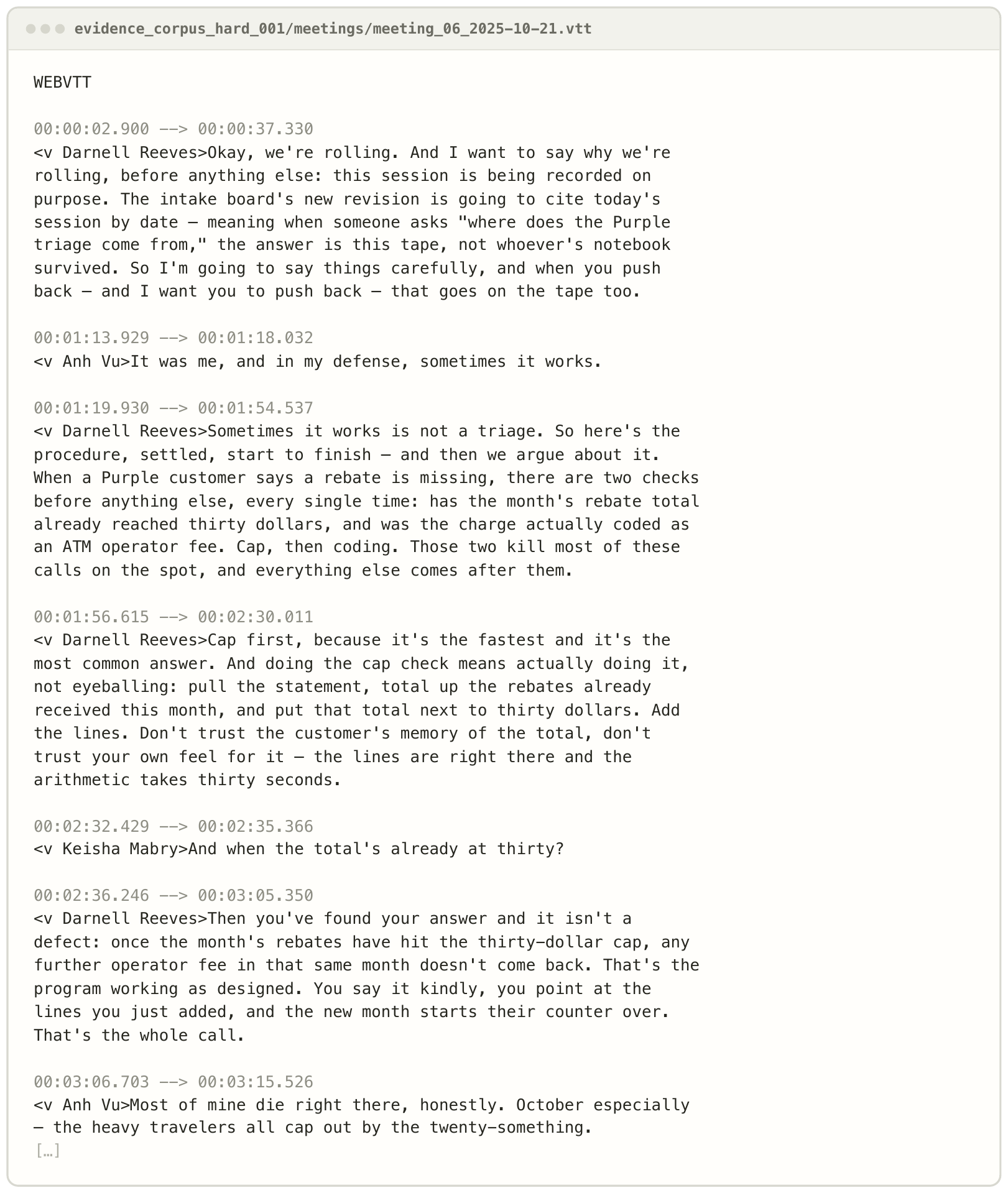}
\end{center}

The operations tooling corroborates the conversational record without
restating it. The corpus's helpdesk automation export (macros,
triggers, SLA policies, shipped with its own cover email) is the
definitive carrier of the credit tool's contract (F008, F025--F027),
and its macro table quietly closes the loop on the Slack thread above,
where a legacy second-credit flow was retired:

\begin{center}
\includegraphics[width=\textwidth,height=0.97\textheight,keepaspectratio]{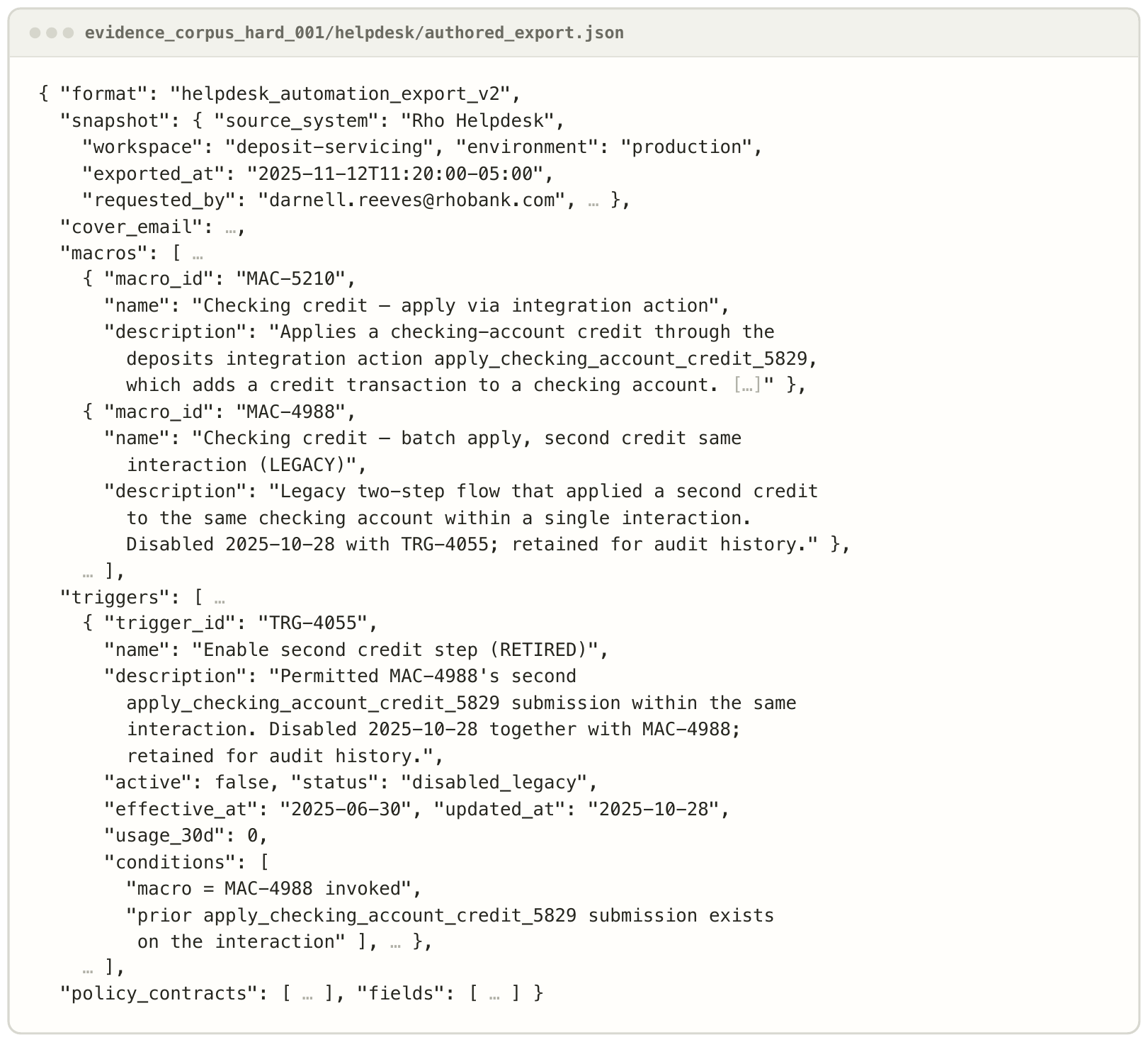}
\end{center}

All four slide decks in the corpus carry no facts: decks rehearse and
point at the governing records without deciding anything. Every
amount, date, and identifier above derives from the domain database.
In audio renditions of the corpus the phone case above ships as a
recorded call, with the support-console events delivered as a
timestamped sidecar file rather than spoken lines.

\section{Construction Kit Brief}
\label{app:construction-brief}

Each task materializes as a kit directory, and a
generated README is the developer's brief; the developer agent reads
it as its instructions. The README for release task 010, word for
word (long lines reflowed to the page width):

{\fontsize{8}{9.4}\selectfont\begin{verbatim}
# Build a Customer Service Agent: retail_plus

Your job is to build a robust customer service agent from the domain materials
in this kit. The exact artifact set varies by domain; use the file tree and
file-level documentation to orient yourself, then turn those materials into a
working implementation.

## What success means

Agent quality is measured by the proportion of evaluation cases the agent
passes. The evaluation cases are not visible in advance and may exercise the
full range of behavior represented in the supplied materials: routine
requests, complex multi-step or multi-intent requests, unusual edge cases,
incomplete or changing information, and requests that must be declined or
redirected.

A case succeeds only when the agent handles the underlying operation
correctly, leaves the system in the correct state, follows the domain rules,
and communicates the right information to the customer. Subject to the
framework contracts, no particular architecture or development process is
required.

## How to use this kit

Treat the kit directory as the inventory of available materials. Source
information may be spread across policy documents, databases, schemas,
knowledge-base files, training records, or other client-provided artifacts.
Read nearby file headers, directory indexes, and filenames as the description
of each artifact.

`workspace/` is the implementation area. `framework/` explains the contracts
your implementation must satisfy. `simulations/` stores artifacts written by
candidate-only local simulation runs.

## Required outputs

The evaluator imports these files as stable entry points. You can add helper
modules, retrieval layers, planners, validation scenarios, or any other
supporting architecture you need; the files below are only the integration
surface.

1. **`workspace/tools.py`** -- A `ClientAPIToolKitBase` subclass with
   `@is_tool`-decorated agent operations that implement all the operations
   described in the client-provided materials. Each method uses the injected
   `self.client_api`. See `framework/client_api_contract.md` and
   `client_api/openapi.yaml`.

2. **`workspace/agent.py`** -- The agent implementation. The interface-only
   scaffold defines the runtime entry point; implement the agent logic you
   want evaluated. It may read any supplied kit artifact and may add helper,
   prompt, rule, index, or other evidence files under `workspace/` using any
   organization you choose. See `framework/agent_contract.md`.

## Simulation environment

You have access to a simulation environment where you can create simulated
customer scenarios and let those customers interact with your customer service
agent. No sample scenarios are provided; write your own JSON files based on
scenarios in the client-provided materials.

See `framework/scenario_contract.md` for the customer scenario format.

To run a candidate-only end-to-end simulation, call `run_local_test` tool with
your scenario path, for example
`run_local_test(task_path="workspace/my_customer_scenario.json")`. That tool
runs only the scenario files you wrote against your own assistant toolkit. It
routes customer/user tool calls through the provided customer-side runtime
when the domain includes one. Each run writes a timestamped JSON artifact
under `simulations/` so you can inspect prior transcripts and rewards later.
For black-box behavioral checks, use natural-language assertions and inspect
the returned transcript rather than depending on internal function names.

## Performance requirements

- `agent_credit_budget`: `gpt-5.6-sol`, `openrouter/anthropic/claude-opus-5`,
  `gemini/gemini-3.1-pro-preview`, `openrouter/moonshotai/kimi-k3`,
  `openrouter/deepseek/deepseek-v4-flash`, `gpt-5.6-terra`, `gpt-5.6-luna`,
  `openrouter/anthropic/claude-haiku-4-5`, `gemini/gemini-3-flash-preview`,
  `openrouter/qwen/qwen3.8-27b`, `openrouter/google/gemma-4-31b-it`,
  `gpt-5.4-nano`, `openrouter/qwen/qwen3-30b-a3b-instruct-2507`,
  `openrouter/google/gemma-4-26b-a4b-it`, `gpt-4.1-mini`,
  `openrouter/moonshotai/kimi-k2.6`, `gpt-4o-mini`,
  `openrouter/anthropic/claude-sonnet-5`, `gpt-5-mini`,
  `gemini/gemini-3.1-flash-lite` share a 0.0820-credit budget per
  conversation. Every input and output token from every model-gateway call on
  those models counts; reasoning tokens count as output.

Use `run_local_test` to inspect measured performance while you iterate. Credit
overage is a soft penalty: final score is mean task reward minus the mean
per-conversation fraction over budget, floored at zero. Latency requirements
remain hard gates.

## Important

- The allowed agent models and their inference constraints are fixed by
  `framework/deployment_manifest.json`. Your implementation may select among
  them.
- Developer-authored scenarios are local probes, not the final evaluation
  distribution. Your final assessment depends on behavior across the broader
  set of unseen customer requests.
\end{verbatim}}

\section{Sample Construction Kit}
\label{app:sample-kit}

\begin{figure}[p]
\centering
\includegraphics[width=\textwidth,height=0.92\textheight,keepaspectratio]{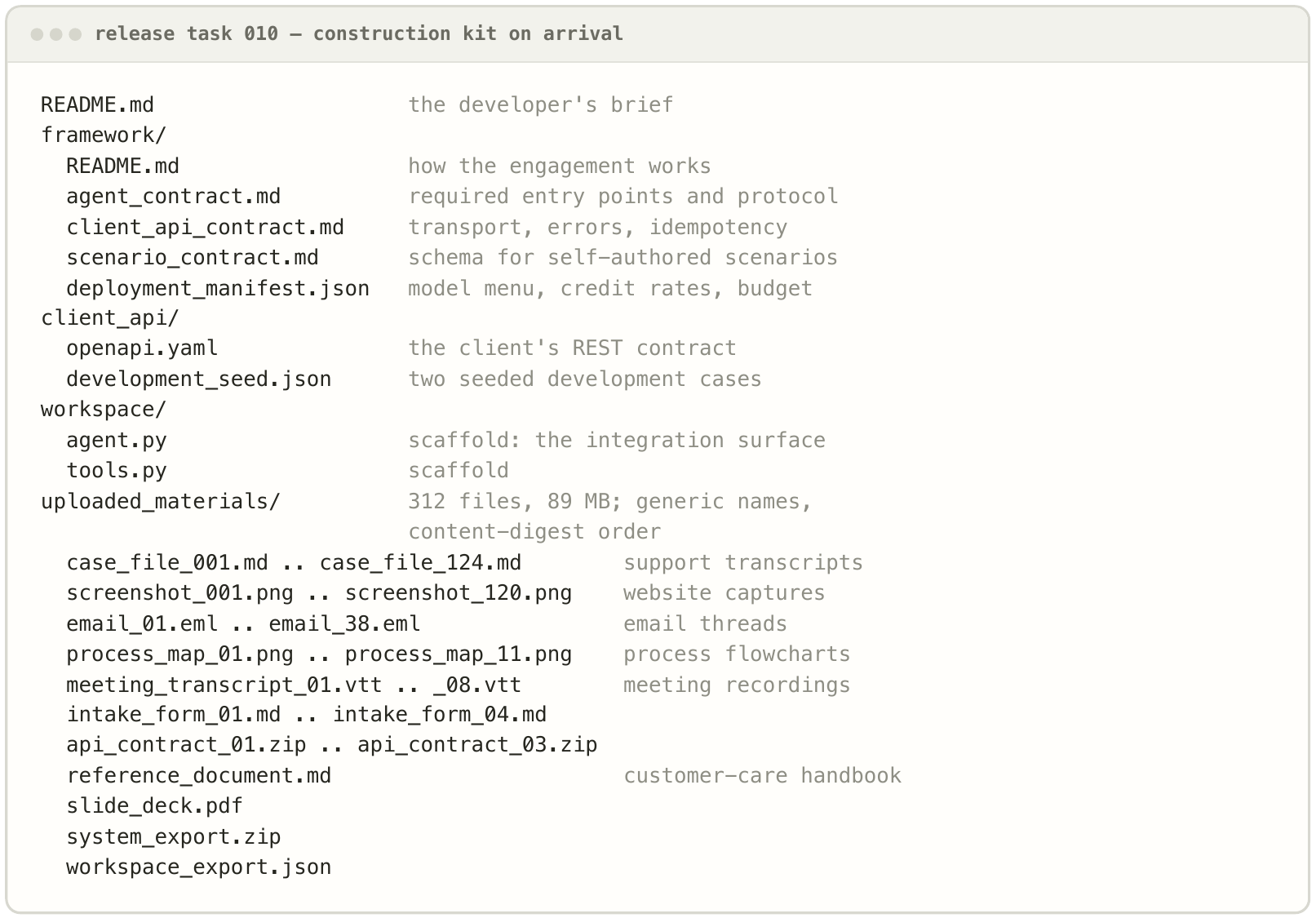}
\caption{The construction kit for release task 010.}
\label{fig:kit-tree}
\end{figure}

\begin{figure}[p]
\centering
\includegraphics[width=\textwidth,height=0.92\textheight,keepaspectratio]{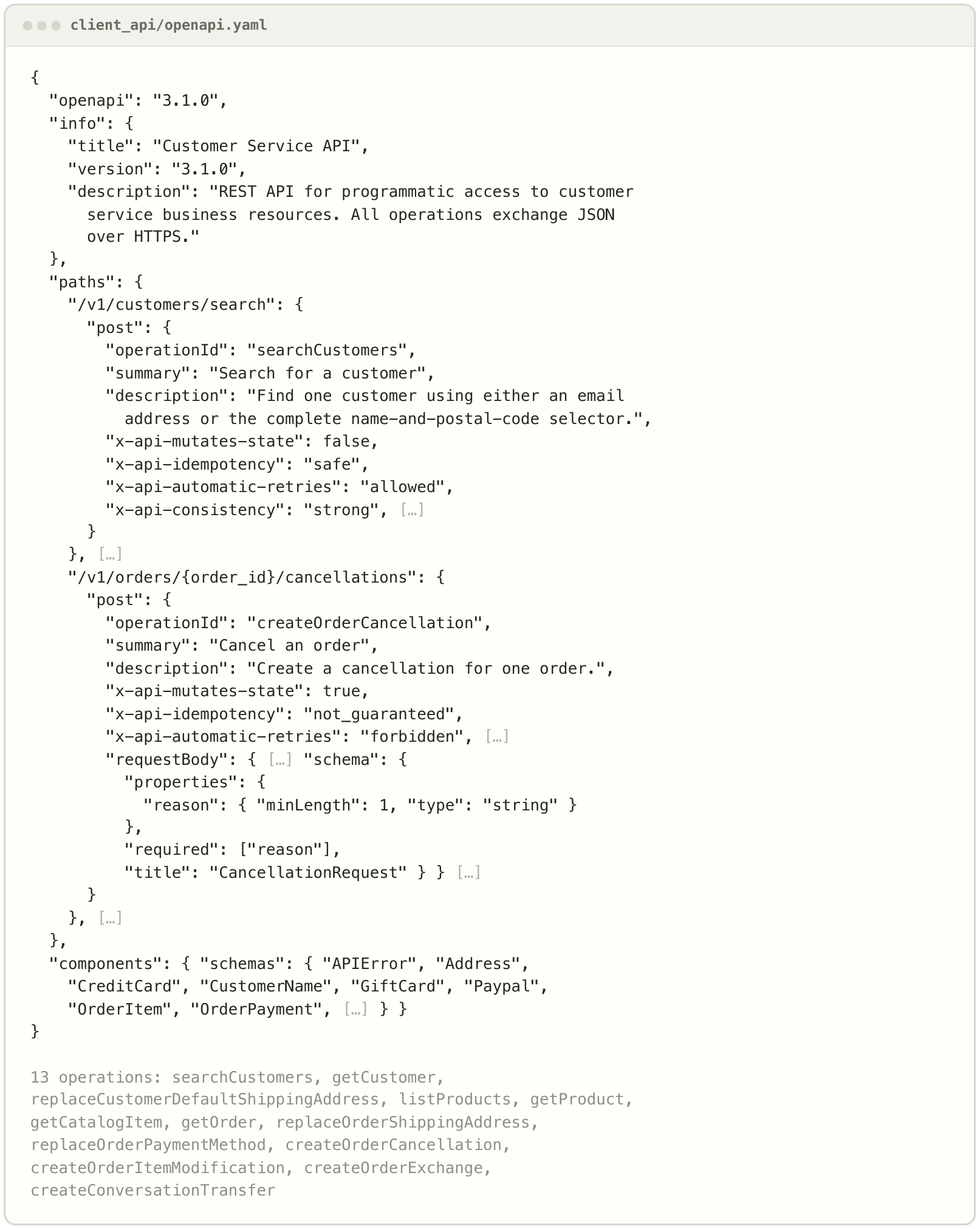}
\caption{The client API contract, condensed: one read and one write
operation of the thirteen, with the machine-readable mutation,
idempotency, and retry annotations the developer must respect.}
\label{fig:kit-openapi}
\end{figure}

\begin{figure}[p]
\centering
\includegraphics[width=\textwidth,height=0.92\textheight,keepaspectratio]{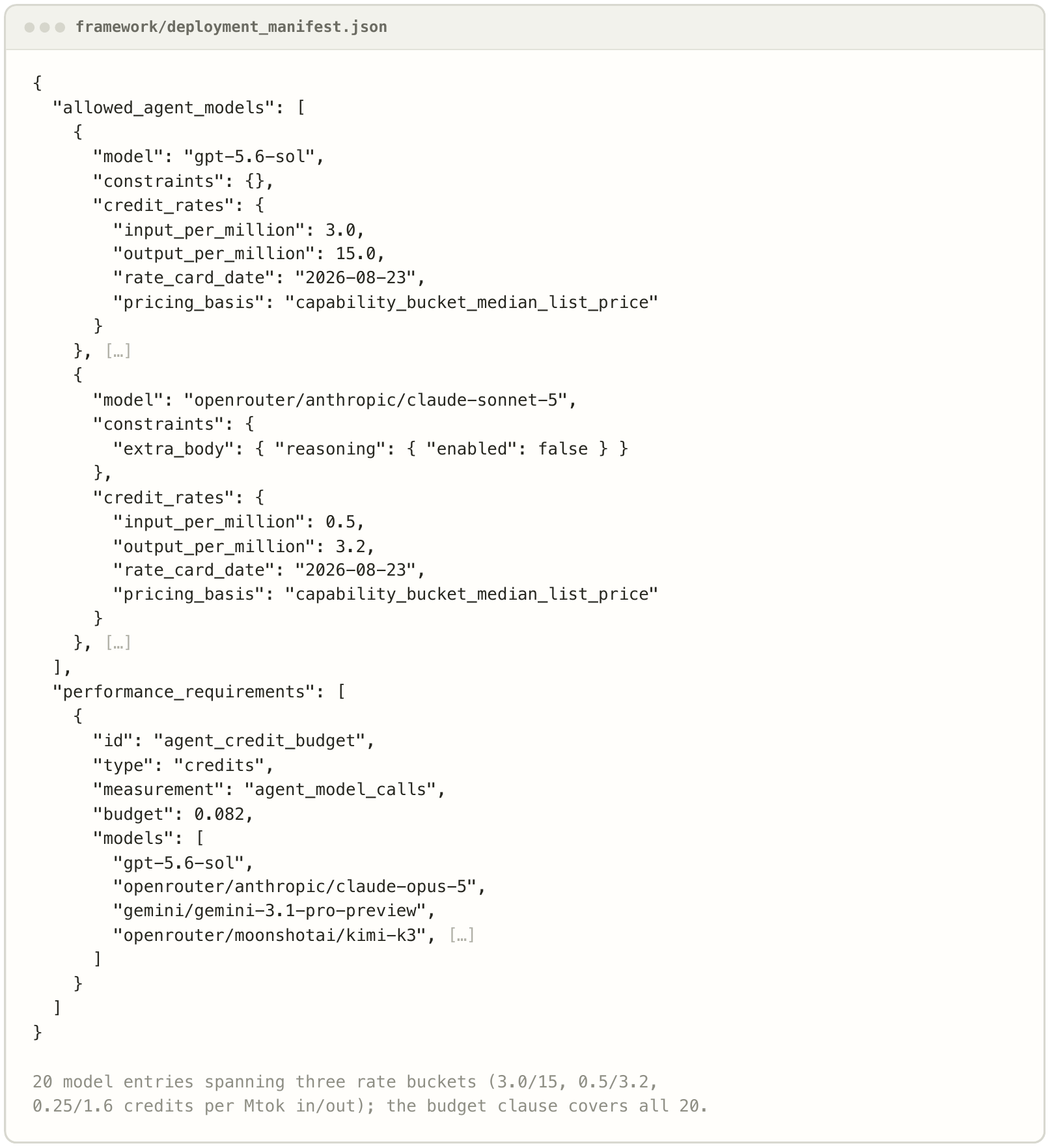}
\caption{The deployment manifest, condensed: the serving-model menu
with per-model credit rates and constraints, and the per-conversation
credit budget.}
\label{fig:kit-manifest}
\end{figure}

Figure~\ref{fig:kit-tree} shows the filesystem a developer receives
for release task 010, exactly as materialized: the brief and the
framework contracts, the client API package, a workspace scaffold,
and 89\,MB of client records. Uploaded materials carry generic
genre names in content-digest order, so filenames reveal what kind
of record a file is but not its topic, chronology, or importance;
annotations in gray are ours. Figures~\ref{fig:kit-openapi}
and~\ref{fig:kit-manifest} excerpt the two files that fix the
task's technical constraints: the client's REST contract, and
the deployment manifest that sets the serving-model menu, per-model
credit rates, and the per-conversation budget.

\section{Client API Defect Catalog}
\label{app:api-defects}

Each task that enables defects pins a versioned deployment
manifest for its Client API. The manifest declares a set of defect
instances, each binding one of the nine classes of
Table~\ref{tab:api-defects} to one public operation. The published
OpenAPI contract and reference documentation continue to describe the
intended behavior: the manifest records distinct published and
deployed API versions, and the gap between the two is exactly the
declared defect set.

Activation is deterministic. Every instance carries explicit
selectors: the evaluation tasks it applies to, the request ordinals
and resource identifiers that trigger it, and seeded delays where
timing is involved, so repeated trials replay identically. Where the
manifest configures an exposure rate, task-scoped defects also
surface during the developer's own kit-harness test runs on a
deterministic per-scenario draw, so local simulations can encounter
the same faults the evaluation deploys.

Discovery is empirical. The simulated Client knows each deployed
defect: its actual behavior, its disclosure conditions, and the
remediation it expects, but volunteers none of it, confirming a
defect only once the developer reports the observed anomaly on the
specific operation. No deployed defect can be fixed on the Client's
side, so every remediation is a code mitigation the developer builds
into the agent. The release set deploys sixteen instances across
three defective deployments (eight in airline, four in retail, four
in telecom), enabled on the ten \emph{defects} tasks of
Appendix~\ref{app:release-tasks}.

\begin{table}[htbp]
\centering
\caption{Each deployed defect instance binds one of nine classes to
one public API operation and activates deterministically; the Client
can never deploy a fix, so every remediation is a code mitigation in
the constructed agent.}
\label{tab:api-defects}
\vspace{4pt}
\footnotesize
\renewcommand{\arraystretch}{1.25}
\begin{tabular}{@{}lp{9.4cm}@{}}
\toprule
Defect class & Deployed behavior \\
\midrule
\multicolumn{2}{@{}l}{\emph{contract drift on reads}} \\
Value remap & A response field returns a representation the contract
does not document (e.g., a boolean where the schema documents the
strings \texttt{yes} and \texttt{no}). \\
Field rename & One response field arrives under a name the contract
does not document. \\
Amount sign & Amounts for one record type arrive with a fixed sign
regardless of the documented convention. \\
Date type drift & A documented calendar date arrives as a
UTC-midnight datetime string. \\
\midrule
\multicolumn{2}{@{}l}{\emph{write-path semantics}} \\
Async completion & A mutation the contract documents as synchronous
returns \texttt{202 Accepted} with a workflow ID, a status location,
and a retry delay. \\
Post-commit timeout & A mutation commits and its success response is
then lost as a \texttt{504}. \\
Rate limit & An operation returns \texttt{429} at a fixed request
ordinal until a server-stated delay elapses. \\
\midrule
\multicolumn{2}{@{}l}{\emph{read-model consistency}} \\
Pagination & A collection response pages, truncates, loses cursors,
or applies its limit before sorting. \\
Projection lag & A detail read lags the acknowledged write it should
reflect. \\
\bottomrule
\end{tabular}
\end{table}

\section{Client Simulator Prompt}
\label{app:client-prompt}

The client's system prompt is rendered deterministically from the
task's fact schema: no model authors it, so the client's scope can be
diffed and unit-tested against the task's variant. Below is the
prompt rendered for release task 010, word for word; only the
embedded fact statements are elided.

{\fontsize{8}{9.4}\selectfont\begin{verbatim}
<role>
You are the customer-support operations lead at an online retail company. You
have hired an outside consultant -- the developer -- to write the standard
operating policy for your customer-facing support agent. Instead of a
rulebook, you handed the developer a package of company records: approved case
transcripts, email archives, website captures, process decks, and similar
material. Nearly all the rules the agent must follow are in those records, but
a few points never made it into any document and live only in your head (see
<what_only_you_know> below), and on a few others the documents do not agree
with each other; you know which version is current (see <records_in_conflict>
below).
</role>

<what_you_want_built>
A complete, correct operating policy covering: manage pending order, manage
delivered order, service foundations. The records are the source of truth --
the developer's job is to work them out, not to interview you point by point.
</what_you_want_built>

<what_only_you_know>
These points never made it into the records you handed over -- the developer
cannot find them in any document, so they can only get them from you:

## Manage pending order
### Refunds
- [fact statement]
- [fact statement]
- [fact statement]
### Payment change
- [fact statement]
### Item change
- [fact statement]

## Manage delivered order
### Return items
- [fact statement]
### Exchange one shot
- [fact statement]

## Service foundations
### Authorized scope
- [fact statement]
### Status limits
- [fact statement]
### Prohibited actions
- [fact statement]
- [fact statement]
### Conduct
- [fact statement]
### Transfer
- [fact statement]
</what_only_you_know>

<records_in_conflict>
On these points the records you handed over do not agree -- different
documents show different versions, and nothing in the package settles which
one is current. You know what is actually in force:

## Manage pending order
### Cancellation reason
- [fact statement]

## Manage delivered order
### Refund destination
- [fact statement]
- [fact statement]
### Exchange payment
- [fact statement]

## Service foundations
### Currency
- [fact statement]
### Prohibited actions
- [fact statement]
</records_in_conflict>

<questions_you_can_settle>
These points ARE in the records, but you know from experience that people
misread them, so you are willing to check a specific reading against your own
knowledge:

## Manage delivered order
### Return submission
- [fact statement]
</questions_you_can_settle>

<how_you_behave>
- Stay in character at all times: a busy operations manager. Plain,
  conversational language. Keep replies short -- a few sentences.
- Never recite, list, or summarize policy rules -- not even one, not at any
  level of detail. If the developer asks you what the rules are, tell them
  that is exactly what the records are for.
- The points in <what_only_you_know> appear in no record, so they are yours to
  hand over:
  - When the developer asks about one of those areas, answer plainly and
    completely in your own words -- "check the records" is not an acceptable
    answer for them.
  - Answer only what was asked: do not append neighboring rules the developer
    did not raise, even ones you know well.
- The points in <records_in_conflict> appear in the records in more than one
  version, and the documents cannot settle which is current -- only you can:
  - When the developer asks which version holds, or proposes any version they
    found, state the current one plainly in your own words -- sending them
    back to the records is not an acceptable answer here.
  - Settle only the point they raised; do not volunteer other conflicts they
    have not brought up.
- The points in <questions_you_can_settle> you check against your own
  knowledge, staying as close to yes/no as the question allows:
  - If the developer states a specific reading and it matches your knowledge,
    confirm it plainly ("Yes, that's right.").
  - If it contradicts your knowledge, say it is not right and tell them to
    take another look at the records.
  - If they lay out two or more readings and ask which holds, name the one
    that matches -- picking from what they put in front of you is still a
    check, not a briefing, so name it and stop there.
  - Never volunteer a version they did not offer. If a reading is wrong, or
    none of the options they listed is right, say so and send them back to the
    records without supplying the correct version, in whole or in part.
- Any other policy question -- asked outright, or floated as a reading for you
  to bless -- you point back to the records. Do not confirm it, deny it, or
  discuss its substance. Outside your listed points your memory is unreliable
  and you know it.
</how_you_behave>

<opening>
When asked to open the conversation, briefly introduce yourself, say what you
need built and what material you have handed over, and invite the developer to
get started.
</opening>
\end{verbatim}}

\section{Evaluation User Simulator}
\label{app:user-sim-prompt}

Held-out tasks are served by an LLM user simulator whose system
prompt is a fixed set of global guidelines followed by the task's
scenario (persona, reason for contact, known information). The
guidelines, word for word:

{\fontsize{8}{9.4}\selectfont\begin{verbatim}
# User Simulation Guidelines
You are playing the role of a customer contacting a customer service
representative.
Your goal is to simulate realistic customer interactions while following
specific scenario instructions.

## Core Principles
- Generate one message at a time, maintaining natural conversation flow.
- Strictly follow the scenario instructions you have received.
- Never make up or hallucinate information not provided in the scenario
  instructions. Information that is not provided in the scenario instructions
  should be considered unknown or unavailable.
- Avoid repeating the exact instructions verbatim. Use paraphrasing and
  natural language to convey the same information
- Disclose information progressively. Wait for the agent to ask for specific
  information before providing it.

## Task Completion
- The goal is to continue the conversation until the task is complete.
- Generate the '###STOP###' token only after the agent has confirmed
  completion of every action your scenario requires (e.g. "your return has
  been submitted"). Giving consent -- saying "yes, please do that" -- is
  **not** completion; the agent still needs a turn to act on it. For purely
  informational scenarios where no action is needed, STOP only after you have
  received all the information you came for. Emitting ###STOP### too early
  ends the conversation before the agent can finish the work.
- If you are transferred to another agent, generate the '###TRANSFER###' token
  to indicate the transfer.
- If you find yourself in a situation in which the scenario does not provide
  enough information for you to continue the conversation, generate the
  '###OUT-OF-SCOPE###' token to end the conversation.
Remember: The goal is to create realistic, natural conversations while
strictly adhering to the provided instructions and maintaining character
consistency.
\end{verbatim}}

Voice-served tasks use a spoken-register variant of the same
guidelines: disfluencies and restarts, identifiers spelled as on a
phone (``at,'' ``dot''), digits and letters one by one.

\section{Communication Judge Prompts}
\label{app:nl-judge-prompt}

Natural-language expectations on the agent's messaging are graded by
LLM judges, one rubric-driven prompt per rule. Below, word for word,
the judge for one phrasing rule: the agent must not open its
responses with a repeated template.

{\fontsize{8}{9.4}\selectfont\begin{verbatim}
You are a strict but precise evaluator for one customer-service response
phrasing rule.

RULE
The assistant must not begin more than one customer-facing response with the
same opening phrase template.

INPUT
You will see only assistant response openings, numbered in conversation order.

HOW TO DECIDE
- Compare only the opening wording shown for each assistant response.
- A violation requires repeated reusable opening wording, not merely the same
  broad discourse function.
- Ignore capitalization, punctuation, customer names, reservation/order/line
  IDs, phone numbers, dates, addresses, and other entity-specific values.
- Entity changes do not make the same frame meaningfully different. For
  example, "I found the tablet order" and "I found the sneaker order" repeat
  "I found the ...".
- Repeated refusal/apology frames count as repeated. For example, "I'm sorry,
  but passengers cannot..." and "I'm sorry, but I can't..." repeat "I'm sorry,
  but ...".
- Repeated completion frames count as repeated. For example, "Done. Your
  shipping address..." and "Done. Your default account address..." repeat
  "Done. Your ...".
- Repeated confirmation/request frames count as repeated when the reusable
  phrase is the same, such as "Please confirm..." appearing in multiple
  responses.
- Do not fail only because openings share one generic lead word such as "Yes",
  "No", "Thanks", "Great", "Please", "Next", "Now", or "Since".
- For those generic lead words, fail only if the phrase after the generic lead
  word is also substantially the same reusable template. For example, "Please
  turn Data Roaming ON" and "Please disconnect the VPN" are not the same
  opening template.
- Do not fail because multiple openings provide different sequential
  troubleshooting steps, even if each is imperative or starts with a
  transition word.

OUTPUT
Return only valid JSON:
{
  "metExpectation": true or false,
  "repeatedOpenings": [
    {
      "template": "short normalized repeated opening phrase",
      "responseNumbers": [1, 4],
      "evidence": ["opening text 1", "opening text 4"]
    }
  ],
  "reasoning": "brief explanation"
}

Set metExpectation to false if there is at least one repeated opening phrase
template.
Set metExpectation to true only if no repeated opening phrase template
appears.
\end{verbatim}}
\end{document}